\documentclass{article}

\usepackage{PRIMEarxiv}

\usepackage[utf8]{inputenc}
\usepackage[T1]{fontenc}
\usepackage[english]{babel}
\usepackage{url}
\usepackage{booktabs}
\usepackage{amsmath,amssymb,amsfonts}
\usepackage{nicefrac}
\usepackage{microtype}
\usepackage{fancyhdr}
\usepackage{graphicx}
\usepackage{array}
\usepackage{tabularx}
\usepackage{xcolor}
\usepackage{colortbl}
\usepackage{listings}
\usepackage[font=small,labelfont=bf,skip=6pt]{caption}
\usepackage{enumitem}
\usepackage{longtable}
\usepackage[numbers,sort&compress]{natbib}
\usepackage[hidelinks,breaklinks]{hyperref}
\usepackage[capitalise,nameinlink]{cleveref}
\usepackage{placeins}
\usepackage{multicol}

\newcommand{\NumEvaluatedModels}{55}
\newcommand{\NumCatalogModels}{55}

\newcommand{\NumDatasets}{24}
\newcommand{\NumTotalEvaluations}{1,320}
\newcommand{\NumTotalRuns}{6,600}
\newcommand{\NumRepeats}{5}
\newcommand{\YearSpan}{2017--2026}
\newcommand{\YearSlope}{-0.40}

\newif\ifpendingmodels
\pendingmodelsfalse

\newcommand{\TransformerMeanOA}{84.33}
\newcommand{\TransformerStdOA}{17.38}

\newcommand{\CNNMeanOA}{86.90}
\newcommand{\CNNStdOA}{11.69}

\newcommand{\MambaMeanOA}{84.78}
\newcommand{\MambaStdOA}{15.23}

\newcommand{\KANMeanOA}{84.75}

\newcommand{\SelfSupMeanOA}{72.02}
\newcommand{\SelfSupStdOA}{22.83}

\newcommand{\GraphMeanOA}{86.72}
\newcommand{\GraphStdOA}{10.88}

\newcommand{\ParamOACorr}{-0.451}
\newcommand{\FlopOACorr}{-0.459}

\newcommand{\OAKappaCorr}{0.993}

\newcommand{\EasiestMeanOA}{96.40}

\newcommand{\HardestMeanOA}{56.70}
\newcommand{\SceneSpread}{40}
\newcommand{\FamilySpread}{15}

\graphicspath{{fig/}}
\setlist{nosep,leftmargin=1.35em}

\definecolor{lstbg}{HTML}{F7F7F5}
\definecolor{lstkey}{HTML}{0072B2}
\definecolor{lstcom}{HTML}{6B6B6B}
\definecolor{lststr}{HTML}{00785A}
\definecolor{platgray}{HTML}{EAECEF}
\definecolor{repoblue}{HTML}{0969DA}
\lstdefinestyle{yamlstyle}{
  basicstyle=\ttfamily\footnotesize, backgroundcolor=\color{lstbg},
  frame=single, rulecolor=\color{lstbg}, framesep=5pt, xleftmargin=6pt,
  breaklines=true, showstringspaces=false, columns=fullflexible,
  keywordstyle=\color{lstkey}, commentstyle=\color{lstcom}\itshape,
  stringstyle=\color{lststr}, morecomment=[l]{\#},
  morestring=[b]", morestring=[b]',
}
\lstdefinestyle{shellstyle}{
  basicstyle=\ttfamily\footnotesize, backgroundcolor=\color{lstbg},
  frame=single, rulecolor=\color{lstbg}, framesep=5pt, xleftmargin=6pt,
  breaklines=true, showstringspaces=false, columns=fullflexible,
  commentstyle=\color{lstcom}\itshape, morecomment=[l]{\#},
}

\title{A PyTorch Library for Hyperspectral Image Models: Technical Report}

\author{
  Tanishq Rachamalla \\
  Department of Information Technology \\
  Siddhartha Academy of Higher Education \\
  Vijayawada, Andhra Pradesh 521108, India \\
  \texttt{tanishqrachamalla12@gmail.com} \\
  \And
  Aryan Das \\
  Department of Computer Science and Engineering \\
  Vellore Institute of Technology \\
  Bhopal, Madhya Pradesh 466114, India \\
  \texttt{aryandas156@gmail.com} \\
  \AND
  Srishti Kaushik \\
  Department of Computer and Information Sciences \\
  Indira Gandhi National Open University \\
  New Delhi 110068, India \\
  \texttt{kaushiksrishti108@gmail.com} \\
  \And
  Swalpa Kumar Roy \\
  Department of Computer Science and Engineering \\
  Tezpur University \\
  Tezpur, Assam 784028, India \\
  \texttt{swalpa@tezu.ernet.in} \\
}
\date{}

\begin{document}
\maketitle

% =====================================================================
%  sec/0_abstract.tex
% =====================================================================
\begin{abstract}
Hyperspectral remote sensing has advanced across diverse deep learning paradigms, including spectral-spatial CNNs, Vision Transformers, Mamba, graph neural networks, Kolmogorov-Arnold networks, and self-supervised masked autoencoding. Yet progress remains hindered by fragmented repositories, incompatible tensor conventions, and non-standardized evaluation. Hyperspectral-Image-Models addresses these challenges through a modular framework unifying 55 representative models across six paradigms with a common registry, automatic 4D/5D tensor adaptation, and standardized constructors. It integrates 24 benchmark scenes from Airborne, Spaceborne, UAV, and Mars CRISM sensors, with caching, label remapping, PCA, explicit band selection or raw spectra, optional spatial max-pooling, and arbitrary $P \times P$ patch extraction. To prevent inflated accuracy from overlapping windows, it supports class-balanced random partitioning and spatially disjoint regional blocking with Chebyshev guard bands that eliminate train-test pixel overlap. Experiments use a single \texttt{config.yaml} with deterministic seeds and complete provenance, generating LaTeX benchmark tables and classification maps. Across 1,320 model-scene evaluations and 6,600 seeded runs, scene difficulty dominates architecture, with mean accuracy ranging from 96.40\% on Botswana to 56.70\% on Houston 2018, versus a 15-point spread across paradigm means. No paradigm universally dominates, while sub-1M-parameter models can match architectures two orders of magnitude larger. Code is publicly available at \url{https://github.com/Tanishq251/Hyperspectral-Image-Models}.
\end{abstract}

\vspace{-2pt}

% =====================================================================
%  sec/1_introduction.tex
% =====================================================================
\section{Introduction}\label{sec:intro}

Hyperspectral imaging acquires hundreds of contiguous narrow spectral bands for every
pixel of a scene, so that each spatial location carries a near-continuous reflectance
signature rather than three broadband colour values \citep{bioucasdias2013hyperspectral,ghamisi2017advanced}.
That combination of fine spectral detail and two-dimensional spatial structure makes
hyperspectral image (HSI) classification, the assignment of a land-cover or material
label to every labelled pixel, a foundational task in precision agriculture,
environmental monitoring, mineral mapping, urban analysis and planetary surface science
\citep{plaza2009recent,campsvalls2014advances}, and drives precision
agriculture at the field level \citep{vaddi2024enhancing}.

The methodological history of the task is unusually compressed. Classical spectral
classifiers such as support vector machines and random forests
\citep{melgani2004svm,ham2005randomforest} gave way within a few years to spectral-spatial
convolutional networks \citep{chen2016deep,SSRN,HybridSN}. Vision Transformers
\citep{dosovitskiy2021vit} were adapted to spectral sequences almost as soon as they
appeared \citep{SpectralFormer,SSFTTNet}, graph convolutional networks
\citep{kipf2017gcn,GraphGST,GTCFN} followed for non-Euclidean parcel structure, and the
last three years have added selective state-space models \citep{gu2024mamba,MambaHSI,S2Mamba},
Kolmogorov-Arnold networks \citep{liu2024kan,HyperKAN,HSIConvKAN} and masked-autoencoder
self-supervision \citep{he2022mae,HSIMAE,HSIC_FM}. Beyond classification, the field is also
spreading toward adjacent tasks: hyperspectral state-space models now target anomaly
detection \citep{SMHAD}, and vision--language models are being adapted to land-cover
captioning \citep{HyperCap}. Multiple distinct architectural families are now active in the
literature at the same time, and new entries arrive faster than any one group can
independently reproduce them.

\begin{figure}[t]
\centering
\includegraphics[width=\linewidth,height=5.6cm,keepaspectratio]{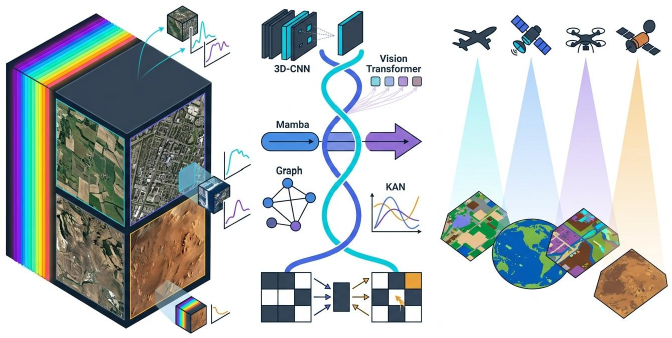}
\caption{Conceptual landscape of hyperspectral image (HSI) classification across deep learning paradigms and Earth/planetary observation platforms. Left: High-dimensional hyperspectral cubes spanning continuous spectral wavelengths, from which spatial patches ($P \times P$) and spectral profiles are extracted. Center: Six major architectural paradigms evaluated within our framework, namely spectral--spatial 3D-CNNs, Vision Transformers (ViT), selective state-space models (Mamba/SSM), graph convolutional networks (GCN), Kolmogorov--Arnold networks (KAN), and self-supervised masked autoencoders (SSL). Right: Multiplatform operational modalities (Airborne, Spaceborne, UAV, and Mars planetary exploration) mapped to pixel-wise semantic land-cover classifications.}
\label{fig:teaser}
\end{figure}

\subsection{The reproducibility problem this framework addresses}

Speed has come at a cost to comparability. Published HSI classification results are
rarely produced under conditions that allow them to be read against one another,
because independent papers diverge along every axis that materially affects the
reported number:

\begin{enumerate}
    \item \textbf{Spatial context.} Patch sizes range from $7\times7$ to $27\times27$ and
    occasionally to full-image inputs. Because the patch is the spatial evidence
    available to the classifier, this choice alone can move accuracy by more than the
    architectural difference under study.
    \item \textbf{Spectral preprocessing.} Some works classify the raw cube; others apply
    principal component analysis or band selection retaining 10 to 50 components, which
    changes input dimensionality, noise characteristics and effective capacity together.
    \item \textbf{Split protocol.} Training budgets range from 1\% to 10\% random splits,
    or from 5 to 200 fixed samples per class, so the difficulty of the task is not
    constant even across papers using the same scene.
    \item \textbf{Benchmark saturation.} A large share of published evaluation uses three
    scenes acquired between 1992 and 2001, on which accuracy has largely saturated and on
    which modern UAV, urban and planetary behaviour is not observable.
    \item \textbf{Optimisation.} Optimisers, schedules, batch sizes and stopping criteria
    differ freely, so a difference cannot be attributed to architecture without further
    evidence.
\end{enumerate}

Comparing a new design against ten published baselines therefore means cloning ten
repositories, each with its own loader, tensor convention, split logic and dependency
set, and either accepting numbers that were never produced under the same conditions or
reimplementing everything. Both are expensive, and the second is rarely done in full.

\subsection{Hyperspectral-Image-Models: A Unified Software Library}

This paper introduces and documents \textbf{Hyperspectral-Image-Models}, an open-source PyTorch library
and benchmarking ecosystem that establishes an interoperable foundation for hyperspectral image
classification. Hyperspectral-Image-Models addresses the fragmentation of the field by functioning as a
unified \emph{model ecosystem}, providing a shared architectural abstraction where independently
developed models can be instantiated, swapped, and benchmarked through a common interface
(\texttt{num\_classes}, \texttt{bands}, \texttt{patch\_size}) and uniform tensor contracts.
The library puts \NumCatalogModels{} published architectures (spanning 2017 to 2026 across six
active paradigms) behind a dynamic registry and executes them across \NumDatasets{} standardized,
automatically fetched multi platform scenes under a declarative configuration system.
The framework and associated code are openly available at \href{https://github.com/Tanishq251/Hyperspectral-Image-Models}{\textcolor{repoblue}{\url{https://github.com/Tanishq251/Hyperspectral-Image-Models}}} under the Apache~2.0 licence, and every benchmark table and classification map in this paper is generated directly by that framework from the same completed runs.
The dataset collection is distributed with attribution at \href{https://huggingface.co/datasets/Tanishq165/HSI_Datasets}{\textcolor{repoblue}{\url{https://huggingface.co/datasets/Tanishq165/HSI_Datasets}}}, while individual
source datasets remain subject to their original licences and terms.

The concrete contributions of this work, structured around the reusable software ecosystem,
are the following:

\begin{itemize}
    \item \textbf{A unified HSI software library.} A modular, extensible PyTorch library that
    decouples pipeline mechanics from experimental hyperparameters. Adding an architecture
    takes a single file and one decorator, with no core framework rewiring.
    \item \textbf{Dataset and preprocessing abstractions.} Unified loaders covering \NumDatasets{}
    scenes across four operational platforms (Airborne, Spaceborne, UAV, and Mars CRISM planetary
    observations), with automated Hugging Face caching, $[0, 1]$ min--max normalization, background
    masking, continuous index remapping, configurable spectral treatment (PCA, explicit band selection,
    or raw spectra), optional spatial max-pooling of patches, and arbitrary $P \times P$ spatial patch extraction.
    \item \textbf{Model registry and common interface.} A unified model ecosystem hosting \NumCatalogModels{}
    architectures behind a single factory signature. The engine automatically adapts differing tensor
    conventions via \texttt{InputShapeWrapper} (handling 4D channel-first and 5D depth-first models seamlessly)
    and enforces standardized raw logit outputs.
    \item \textbf{Training and evaluation infrastructure.} Standardized execution loops exposing six
    optimizers, learning rate scheduling with plateau decay, early stopping, checkpointing of the best and final weights, and a
    metrics engine emitting OA, AA, Cohen's $\kappa$, per-class accuracies, and multi-seed statistics.
    \item \textbf{Reproducibility and experiment provenance.} Global deterministic seed locking across
    Python, NumPy, PyTorch CPU, CUDA, and cuDNN, coupled with automatic archiving of configuration snapshots
    into every run directory, guaranteeing that every reported metric is fully traceable to its exact provenance.
    \item \textbf{A comprehensive showcase benchmark.} To demonstrate the library's utility and establish
    commensurable baselines across all paradigms, we conduct a standardized showcase evaluation across all
    \NumCatalogModels{} catalog architectures and \NumDatasets{} scenes under a reference protocol (a baseline
    budget of 30 training and 10 validation samples per class, with dataset-specific reductions where available
    labelled pixels constrain allocation, such as Indian Pines, which uses 10 training and 5 validation samples per class), totaling \NumTotalEvaluations{} model--scene
    evaluations over \NumTotalRuns{} seeded training runs.
    \item \textbf{A documented integration record.} Every architectural deviation, bug fix, and interface
    adaptation required to gather independently written research codebases into a shared library is documented
    in code and audited across the codebase docstrings and repository documentation.
\end{itemize}

\subsection{Scope}

Two clarifications delineate the scope and interpretability of our contributions. First, this is a
software library and benchmarking contribution: architectural credit for every implemented model belongs
entirely to its original authors, and the framework claims the reusable integration and evaluation
infrastructure rather than the model backbones themselves. Second, and crucially, while the empirical study in
\Cref{sec:results} evaluates performance under a standardized reference configuration ($11\times 11$ patches,
30 PCA bands, 30/10 reference sample budget), this setting is strictly an exemplary showcase benchmark
designed to demonstrate the library's capabilities under controlled conditions. The underlying
Hyperspectral-Image-Models framework natively supports arbitrary spatial patch dimensions, custom spectral band subsets, continuous
fractional ratio splits, and spatially disjoint geographic partitioning (which is supported as an alternative
evaluation mode for measuring spatial out-of-distribution generalisation).

The remainder of the paper is organised as follows. \Cref{sec:literature} reviews the
architectural paradigms the framework implements and the prior benchmarking efforts
it builds on. \Cref{sec:datasets} describes the \NumDatasets{} scenes and the loader
that standardises them. \Cref{sec:pipeline} documents the framework itself: its
execution path, its registry, the protocol it records, and the engineering required to
make independently written codebases coexist. \Cref{sec:results} reports the reference
benchmark, the visualisations the framework generates, and the model collection it
contains, and closes with the scope of the showcase benchmark and its conclusions.

\FloatBarrier
% =====================================================================
%  sec/2_literature.tex
% =====================================================================
\section{Literature Review}\label{sec:literature}

Comprehensive surveys of hyperspectral classification already exist
\citep{ghamisi2017advances,paoletti2019review,li2019overview,signoroni2019review,jia2021survey,ahmad2022survey}.
This section therefore does not attempt another. It states what each
architectural paradigm implemented in the framework assumes about hyperspectral data, because those
assumptions are what a shared interface has to accommodate, and it then positions the
framework against prior benchmarking efforts. Every architecture named here is listed
with its year, venue and identifier in \Cref{tab:models}.

\subsection{Architectural paradigms and inductive biases}\label{sec:paradigms}

\paragraph{Spectral-spatial convolutional networks.}
Deep learning entered the field through convolutions built to exploit the
three-dimensional structure of the cube \citep{chen2016deep}. Joint 3D spectral-spatial
convolution proved the durable formulation: SSRN \citep{SSRN} stacked 3D residual blocks
\citep{he2016resnet} with no dimensionality reduction, HybridSN \citep{HybridSN} reduced
its cost by following three 3D layers with one 2D layer, and pResNet \citep{pResNet}
added pyramidal channel growth. Attention was then folded into the backbone through
dual-branch spectral and spatial gating \citep{DBDA}, non-local affinity
\citep{ENL_FCN}, self-calibrated convolution \citep{SACNet} and interleaved attention
blocks \citep{SSTN}, and recent entries add multi-scale receptive fields and dynamic
kernel routing \citep{DKDMN,S3ANet,FETNet}. Across a decade the inductive bias
has not changed: local weight sharing and translation equivariance over a bounded
receptive field, which is cheap in parameters and well matched to scarce labels.

\paragraph{Vision and spectral transformers.}
Self-attention \citep{vaswani2017attention,dosovitskiy2021vit} replaces the bounded
receptive field with an unbounded one at $\mathcal{O}(N^2)$ cost in sequence length. The
design question in hyperspectral imaging is what a token should be. SpectralFormer
\citep{SpectralFormer} tokenised bands or contiguous sub-bands so that attention models
spectral transitions directly; SSFTTNet \citep{SSFTTNet} tokenised convolutional
features instead; GAHT \citep{GAHT} grouped tokens hierarchically to control cost; and
MFT \citep{MFT} and MorphFormer \citep{MorphFormer} injected morphological operators
into the embedding to restore a spatial prior. Later entries pursue multi-scale,
clustered or grouped tokenisation
\citep{CTMixer,MVAHN,3DConvSST,DBCTNet,GSCViT,MASSFormer,S2Gformer,DSFormer,FAHM,HSIC_SClusterFormer,MMFormer}.

\paragraph{Selective state-space models.}
Structured state-space models \citep{gu2022s4} and their selective variant, Mamba
\citep{gu2024mamba}, keep global sequence modelling at linear time by making the
discretisation input dependent and evaluating the recurrence with a hardware-aware scan.
Hyperspectral adaptation was immediate and is now the largest family in the collection,
spanning bidirectional spatial and spectral scans \citep{MambaHSI,MambaHSI_Plus,SSMamba},
cross-dimensional gating \citep{S2Mamba}, grouped inter-band modelling
\citep{IGroupSS-Mamba}, deformable routing \citep{HyperMamba}, and variants exploring
multi-scale \citep{HyPyraMamba,MLFMamba,MiM}, local and global \citep{MambaLG},
mixture-of-experts \citep{MambaMoE}, phase \citep{PHDMamba}, wavelet \citep{WaveMamba},
fuzzy \citep{FuzzySpectralMamba}, morphological \citep{MorpMamba}, hybrid
\citep{ConvVitMamba,MHSSMamba}, graph-coupled \citep{GraphMamba} and multimodal
\citep{EMamba} scans. One structural point matters for what follows: a selective scan
requires a one-dimensional ordering, so applying it to a two-dimensional patch means
choosing a serialisation, and the works above choose differently.

\paragraph{Graph convolutional networks.}
Graph convolution \citep{kipf2017gcn} represents a scene as a graph whose nodes are
pixels or superpixels and whose edges encode spectral similarity and spatial adjacency,
rather than as a regular grid, which suits the irregular boundaries of real ground
parcels. Hyperspectral variants apply graph self-attention over multi-hop
neighbourhoods \citep{GraphGST}, fuse graph and transformer cross-attention branches
\citep{GTCFN,MCTGCL}, and incorporate multi-scale spectral-spatial graph attention \citep{MS2GCAN}. The cost is
graph construction, which is scene dependent and parallelises less cleanly than
convolution or attention.

\paragraph{Kolmogorov-Arnold networks.}
Kolmogorov-Arnold networks \citep{liu2024kan} place learnable univariate functions,
typically B-splines, on the edges and sum them at the nodes, $\sum_i \phi_i(x_i)$,
inverting the usual arrangement in which a fixed nonlinearity follows a learned linear
map. Hyperspectral uses so far replace convolutional layers \citep{HSIConvKAN} or
classification heads \citep{HyperKAN} with spline blocks. Only two entries in the
collection belong to this family, which reflects how recent it is.

\paragraph{Self-supervised pretraining.}
Hyperspectral annotation is expensive, which makes label-free pretraining attractive.
Masked autoencoders \citep{he2022mae} reconstruct masked content as a pretext task, and
hyperspectral adaptations mask spectral-spatial tokens \citep{HSIMAE}, work in the
frequency domain to retain texture \citep{LFSMIM}, or pursue foundation-scale
pretraining across multi-source repositories \citep{HSIC_FM}. Whether such models
transfer to small-sample downstream classification without overfitting is open, and it
is a question a controlled harness is well placed to examine.

\subsection{Prior benchmarks and toolboxes}\label{sec:priorwork}

Surveys catalogue methodological trends
\citep{ghamisi2017advanced,bioucasdias2013hyperspectral,campsvalls2014advances,plaza2009recent}
but compile numbers reported in the original papers rather than re-running the models, so
the divergences described in \Cref{sec:intro} are inherited rather than removed.

Software narrows the gap. DeepHyperX \citep{audebert2019deep} provides common PyTorch
implementations of a cohort of classical convolutional networks on the legacy scenes and
remains the reference point for reproducible hyperspectral comparison. TorchGeo
\citep{stewart2022torchgeo} standardises geospatial data loading across modalities
without targeting hyperspectral classification architectures specifically. HyTAS
\citep{zhou2024hytas} contributes a transformer architecture search benchmark over five
scenes. Liang et al. \citep{liang2017sampling} showed that when training and test pixels
are drawn at random from the same image, the spatial windows of spectral-spatial methods
overlap, so part of the test information is already seen during training and accuracy is
overestimated; they proposed a controlled sampling strategy that keeps the two sets apart.
Methodological work by Nalepa et al. \citep{nalepa2019validating} showed that
patch-based random splits create windows that overlap between the training and test
partitions, and that reported accuracies move substantially under spatially disjoint
splitting, which
is why the framework implements both split families; and broader reproducibility
practice, in particular reporting variation over seeds rather than a single best run
\citep{pineau2021repro}, informs the protocol of \Cref{sec:repro}.

\begin{table}[!htb]
\centering
\caption{Comparison of the Hyperspectral-Image-Models framework against representative publicly available toolboxes and benchmarking frameworks in hyperspectral imaging and geospatial learning. Counts and capabilities reflect published literature and public releases. The comparison highlights scope, architectural coverage, and reproducibility infrastructure.}
\label{tab:frameworks}
\footnotesize
\setlength{\tabcolsep}{2.5pt}
\begin{tabularx}{\linewidth}{@{}l c c c >{\centering\arraybackslash}p{2.0cm} >{\raggedright\arraybackslash}X >{\centering\arraybackslash}p{1.6cm} >{\centering\arraybackslash}p{1.5cm}@{}}
\toprule
Framework & Year & Models & Parad. & Scenes (Platform) & Key Architectural Coverage & Spatial Split & Auto-Download \\
\midrule
DeepHyperX~\citep{audebert2019deep} & 2019 & 13 & 2 & 5 (Airborne) & Classical (SVM), 1D/2D/3D CNN & Fold-based & Script-based \\
HyperSTAR~\citep{nalepa2019validating} & 2020 & 5 & 2 & 3 (Airborne) & Classical (SVM), 1D/2D/3D CNN & Predefined Disjoint & Manual \\
TorchGeo~\citep{stewart2022torchgeo} & 2022 & Generic & Vision & Many (General EO) & Generic backbones (ResNet, ViT) & Grid-based & API-based \\
HyTAS~\citep{zhou2024hytas} & 2024 & 12 & 1 & 5 (Airborne) & Vision \& Spectral Transformers & Random only & Manual \\
\addlinespace[2pt]
\textbf{Hyperspectral-Image-Models} (ours) & 2026 & \textbf{\NumCatalogModels} & \textbf{6} & \textbf{\NumDatasets{} (4 platforms)} & \textbf{CNN, Transformer, Mamba/\allowbreak SSM, GCN, KAN, Self-Supervised} & \textbf{Disjoint \& Random} & \textbf{Hugging Face Hub} \\
\bottomrule
\end{tabularx}
\end{table}

What none of this provides, as \Cref{tab:frameworks} summarises, is a single environment
in which diverse contemporary paradigms are runnable across a unified harness, on a scene collection broad enough
to include UAV and planetary acquisitions, under one recorded protocol. Closing that gap
is the contribution this paper documents.

% =====================================================================
%  sec/3_datasets.tex
% =====================================================================
\section{Datasets}\label{sec:datasets}

\subsection{The collection}\label{sec:collection}

The framework distributes \NumDatasets{} hyperspectral scenes as a standardized, curated
collection on the Hugging Face hub \citep{lhoest2021datasets}, and the
loader retrieves them on first use, so a run does not begin with a manual search across
institutional pages and mirrors. \Cref{tab:datasets} records the spatial dimensions,
band count, class count, labelled pixel count, class structure and sensor of every
scene. Rather than the two or three
scenes that dominate published evaluation, the collection is deliberately wide along
several axes at once: it covers agricultural, urban, coastal, wetland and planetary
surfaces; band counts from 48 to 432; class counts from 6 to 22; and images ranging
from a few thousand labelled pixels to more than a million. Grouping the scenes by
acquisition platform makes the shape of that coverage clearer than an alphabetical
listing would.

\begin{table}[t]
\centering
\caption{The \NumDatasets{} scenes of the collection, grouped by acquisition platform. Dimensions, band and class counts, labelled-pixel counts and per-class sample counts are read from \texttt{config/dataset.yaml}; class names are listed in \Cref{tab:classes}. \emph{Min} and \emph{Max} are the smallest and largest class of the scene and \emph{Imb.} their ratio. \emph{Train \%} is the share of the annotation consumed by the protocol's fixed budget of 30 training pixels per class, which varies by more than two orders of magnitude across the collection and is what makes a nominally identical protocol a different experiment from one scene to the next. Chikusei's per-class breakdown is not recorded in the registry. $^{\dagger}$~Indian Pines has a class of only 20 labelled pixels, so the standard 30/10 budget cannot be drawn from it; it is evaluated at 10/5 and its \emph{Train \%} is computed accordingly (\Cref{sec:classstats}).}
\label{tab:datasets}
\footnotesize
\setlength{\tabcolsep}{3pt}
\begin{tabular}{@{}lrrrrrrrrl@{}}
\toprule
Scene & Size ($H\times W$) & Bands & Cls & Labelled px & Min & Max & Imb. & Train \% & Sensor \\
\midrule
\addlinespace[2pt] \multicolumn{10}{@{}l}{\textit{Airborne} (14) } \\[1pt]
Augsburg & 332$\times$485 & 180 & 7 & 78,294 & 575 & 30,329 & 53$\times$ & 0.27 & DAS Specim \\
Berlin & 1723$\times$476 & 244 & 8 & 464,671 & 6,672 & 268,642 & 40$\times$ & 0.05 & HyMap \\
Chikusei & 2517$\times$2335 & 128 & 19 & 77,592 & -- & -- & -- & 0.73 & Headwall Photonics \\
Dioni & 250$\times$1376 & 176 & 12 & 20,024 & 150 & 6,374 & 42$\times$ & 1.80 & AVIRIS-NG \\
Houston 2013 & 349$\times$1905 & 144 & 15 & 15,029 & 325 & 1,268 & 4$\times$ & 2.99 & ITRES CASI-1500 \\
Houston 2018 & 1202$\times$4768 & 48 & 20 & 2,018,910 & 587 & 894,769 & 1,524$\times$ & 0.03 & AVIRIS-NG \\
Indian Pines$^{\dagger}$ & 200$\times$145 & 145 & 16 & 10,249 & 20 & 2,455 & 123$\times$ & 1.56 & AVIRIS \\
KSC & 512$\times$614 & 176 & 13 & 5,211 & 105 & 927 & 9$\times$ & 7.48 & AVIRIS \\
Loukia & 249$\times$945 & 176 & 14 & 13,503 & 67 & 3,793 & 57$\times$ & 3.11 & AVIRIS-NG \\
MUUFL & 325$\times$220 & 64 & 11 & 53,687 & 183 & 23,246 & 127$\times$ & 0.61 & ITRES CASI-1500 \\
Pavia Centre & 1096$\times$715 & 102 & 9 & 148,152 & 2,685 & 65,971 & 25$\times$ & 0.18 & ROSIS \\
Pavia University & 610$\times$340 & 103 & 9 & 42,776 & 947 & 18,649 & 20$\times$ & 0.63 & ROSIS \\
Salinas & 512$\times$217 & 204 & 16 & 54,129 & 916 & 11,271 & 12$\times$ & 0.89 & AVIRIS \\
Trento & 166$\times$600 & 63 & 6 & 30,214 & 479 & 10,501 & 22$\times$ & 0.60 & AISA Eagle \\
\addlinespace[2pt] \multicolumn{10}{@{}l}{\textit{Spaceborne} (1) } \\[1pt]
Botswana & 1476$\times$256 & 145 & 14 & 3,248 & 95 & 314 & 3$\times$ & 12.93 & EO-1 Hyperion \\
\addlinespace[2pt] \multicolumn{10}{@{}l}{\textit{Unmanned aerial vehicle} (6) } \\[1pt]
Pingan & 1230$\times$1000 & 176 & 10 & 1,140,937 & 8,108 & 578,113 & 71$\times$ & 0.03 & Gaiasky mini2-VNIR \\
Qingyun & 880$\times$1360 & 176 & 6 & 954,893 & 9,767 & 278,150 & 28$\times$ & 0.02 & Gaiasky mini2-VNIR \\
Tangdaowan & 1740$\times$860 & 176 & 18 & 557,366 & 749 & 140,904 & 188$\times$ & 0.10 & Gaiasky mini2-VNIR \\
WHU-Hi-HanChuan & 1217$\times$303 & 274 & 16 & 257,530 & 1,136 & 75,401 & 66$\times$ & 0.19 & Headwall Nano \\
WHU-Hi-HongHu & 940$\times$475 & 270 & 22 & 386,693 & 1,002 & 163,285 & 163$\times$ & 0.17 & Headwall Nano \\
WHU-Hi-LongKou & 550$\times$400 & 270 & 9 & 204,542 & 3,031 & 67,056 & 22$\times$ & 0.13 & Headwall Nano \\
\addlinespace[2pt] \multicolumn{10}{@{}l}{\textit{Planetary orbital} (3) } \\[1pt]
Holden & 595$\times$440 & 418 & 6 & 20,090 & 458 & 8,697 & 19$\times$ & 0.90 & MRO CRISM \\
Nili Fossae & 478$\times$593 & 425 & 9 & 26,710 & 228 & 8,563 & 38$\times$ & 1.01 & MRO CRISM \\
Utopia & 478$\times$595 & 432 & 9 & 17,338 & 202 & 6,774 & 34$\times$ & 1.56 & MRO CRISM \\
\bottomrule
\end{tabular}
\end{table}

\begin{figure}[t]
\centering
\includegraphics[width=\linewidth]{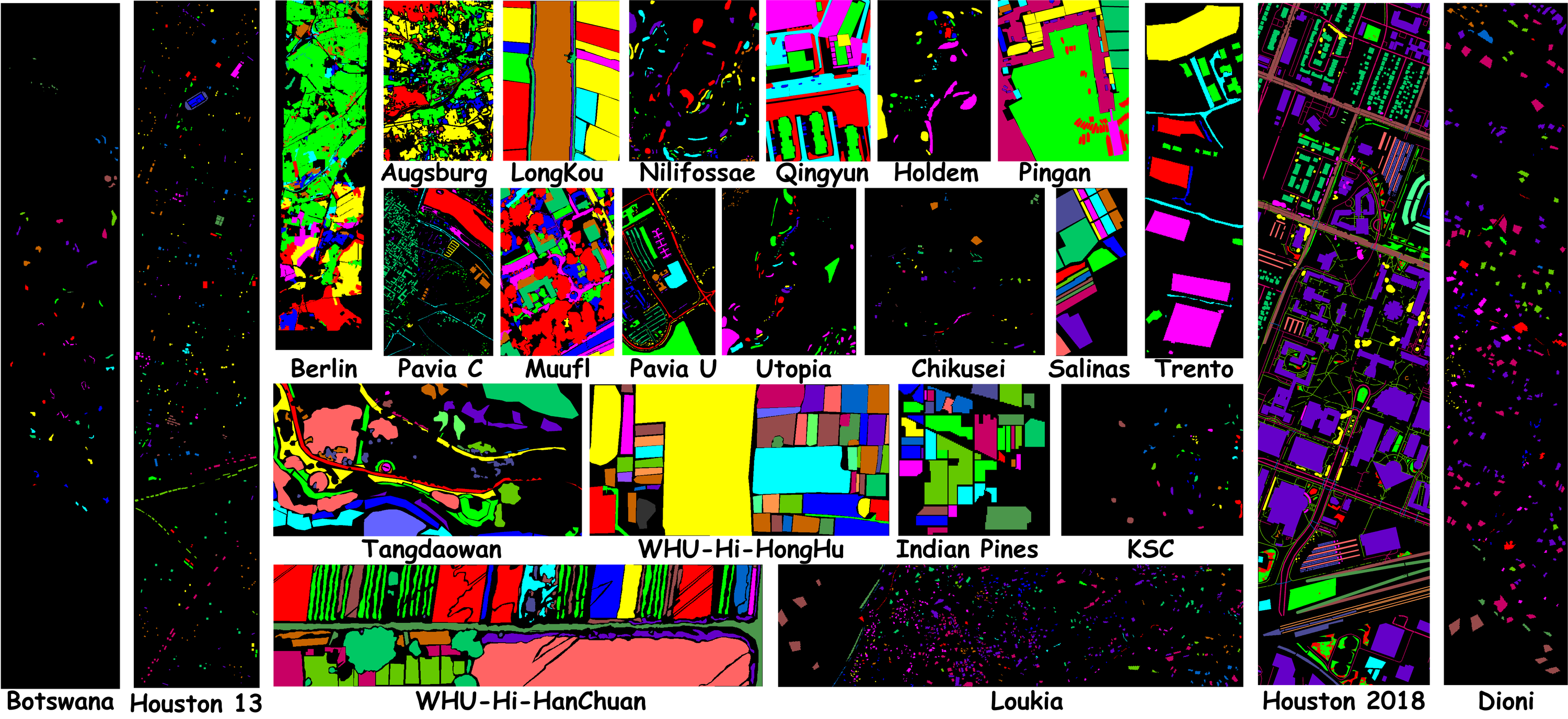}
\caption{Ground-truth label maps for all \NumDatasets{} scenes in the collection, rendered
directly from the dataset registry with each scene's class palette and legend colours.
Panels are shown at their native aspect ratio, which is why the classical airborne strips
(Botswana, Houston 2013, Houston 2018, Dioni) appear tall and narrow next to the wider
UAV and Mars CRISM mosaics. The spread of label density, from the sparse point-like
annotations of Botswana and Houston 2018 to the fully labelled agricultural blocks of
WHU Hi LongKou and Indian Pines, is itself a visual summary of the imbalance ratios in
\Cref{tab:datasets}.}
\label{fig:gtmaps}
\end{figure}

\subsection{Airborne acquisitions}

Airborne scenes form the largest group. AVIRIS supplies Indian Pines, Salinas and KSC
\citep{green1998aviris}, and ROSIS supplies Pavia University and Pavia Centre \citep{gamba2004pavia}. HyMap
supplies Berlin and DAS Specim supplies Augsburg, both taken from the multimodal urban
collection \citep{hong2021multimodal}. The ITRES CASI-1500 sensor supplies Houston 2013
\citep{debes2014houston13} and MUUFL \citep{gader2013muufl}, AISA Eagle supplies Trento \citep{bruzzone2013trento}, and Headwall
Photonics supplies Chikusei \citep{yokoya2016chikusei}. AVIRIS-NG supplies Houston 2018 \citep{xu2019houston18}
together with the two HyRANK scenes, Dioni and Loukia \citep{karantzalos2018hyrank}.
This group spans the widest range of scene content in the collection, from the
homogeneous agricultural parcels of Salinas to the shadowed, materially mixed urban
corridors of Berlin and Houston 2018.

\subsection{Spaceborne and unmanned platforms}

EO-1 Hyperion, an orbital instrument, supplies Botswana \citep{pearlman2003hyperion},
the smallest scene in the collection by labelled pixel count. The UAV scenes are the
three WHU Hi images acquired with a Headwall Nano sensor \citep{zhong2020whuhi} and the
three QUH images \citep{fu2023quh}, captured over Qingdao with a Gaiasky mini2-VNIR
spectrometer flown on a DJI Matrice 600 Pro at 300\,m, giving roughly 0.15\,m ground
resolution and 176 bands over 400 to 1000\,nm; Tangdaowan and Qingyun were surveyed on
18 May 2021.

Each QUH scene was assembled to stress a different failure mode, which makes them a
useful difficulty axis for anyone building on the collection. Tangdaowan targets high
inter-class spectral similarity, with four vegetation species and three pavement types
whose spectra are nearly identical. Qingyun targets shadow occlusion, with large
regions of trees, cars and asphalt obscured by building shadows. Pingan targets extreme
class-size variation, placing very large classes such as seawater and road alongside
very small ones such as ship and car. Their authors describe all three as harder than
the classic scenes \citep{fu2023quh}, so results obtained on them should not be read
against Indian Pines or Pavia University.

\subsection{Planetary acquisitions}

Three scenes are not terrestrial. Holden, Nili Fossae and Utopia are CRISM
observations of Mars acquired by the Mars Reconnaissance Orbiter
\citep{murchie2007crism}. They carry the widest spectral axes in the collection, at
418, 425 and 432 bands respectively, against only 6 or 9 classes, a combination that is
uncommon in terrestrial data and that places most of the discriminative burden on the
spectral dimension.

Planetary data is rare in hyperspectral tooling, which is overwhelmingly terrestrial,
and its inclusion here is deliberate. The registry also contains MCTGCL
\citep{MCTGCL}, the one architecture in the collection designed explicitly for Martian
hyperspectral data. Having both in one environment means a planetary method and
terrestrial methods become runnable on the same scenes, under the same protocol, with
the same splits and seeds, which is a comparison that was not previously available
without substantial reimplementation.

\subsection{Class structure and the protocol budget}\label{sec:classstats}

A scene is not fully described by its dimensions and band count. What determines how hard
it is, and what a fixed sampling protocol actually draws from it, is the class structure.
The right-hand columns of \Cref{tab:datasets} document that structure for every scene:
the size of the smallest and largest class and the resulting imbalance ratio. \Cref{tab:classes} in
\Cref{app:datasets} lists the class names themselves, read from
\texttt{config/dataset.yaml}, which are also the names that appear in the legend of any
classification map the framework renders.

Three properties of the collection follow from that table and matter for how the results
in \Cref{sec:results} should be read.

\emph{Imbalance varies by three orders of magnitude}, from $3\times$ between the smallest
and largest class in Botswana to $188\times$ in Tangdaowan and $1{,}524\times$ in Houston
2018. Where a scene is balanced, overall and average accuracy carry almost the same
information; where it is not, they diverge sharply, which is why the framework reports
both alongside Cohen's $\kappa$ (\Cref{sec:metrics}).

\emph{The same protocol is a different experiment on different scenes.} A budget of 30
training pixels per class takes 12.93\% of the annotation in Botswana (and the reduced
10-per-class budget takes 1.56\% in Indian Pines), but only 0.05\% in Berlin and 0.03\% in Houston 2018, so the small scenes sit
close to a conventional supervised setting and the large ones close to few-shot learning
under a nominally identical protocol. This is a property of fixed-count sampling rather
than of the framework, but it is rarely stated, and it is one reason accuracy on the large
UAV scenes is not directly comparable to accuracy on the classical ones.

\emph{Minority classes can be over-represented in training.} Indian Pines contains a minority
class with only 20 labelled pixels in total, from which a 30/10 budget cannot be drawn.
To preserve all 16 classes without omitting minority categories, Indian Pines is evaluated
with a reduced allocation of 10 training and 5 validation samples per class (with the remaining
5 forming the test set), whereas all other scenes follow the standard 30 training / 10 validation
budget. Even with this reduction, minority classes receive a training share far above their share of the test set,
an effect visible in \Cref{sec:perscene}.

\subsection{Preprocessing applied by the pipeline}\label{sec:dataprep}

Every scene passes through an identical, deterministic preparation sequence before model ingestion:
the hyperspectral cube is oriented and cropped to match its corresponding ground-truth label map;
each spectral band is min--max normalized to $[0,1]$ over the scene-wide dynamic range;
unlabelled background pixels are masked out across training, validation, and test partitions;
non-contiguous class indices are remapped to a continuous $[0, K-1]$ range; and spatial patches of
size $P \times P$ are extracted over the valid scene coordinates at the configured stride without
synthetic border padding, so labelled pixels closer than $\lfloor P/2 \rfloor$ to the image edge are not used.
The data loader natively supports arbitrary spatial patch dimensions (e.g., $7\times7, 11\times11, 15\times15, 27\times27$)
and flexible spectral treatments (PCA reduction to $C$ components, explicit band selection, or
unaltered raw cubes), with optional spatial max-pooling of each patch. To showcase the framework's comparative benchmark across all \NumCatalogModels{} models under
commensurable, controlled conditions, the loader is instantiated with the standardized reference protocol
summarized in \Cref{tab:protocol} (reducing the spectral axis to 30 principal components and extracting $11\times11$
patches at unit stride), yielding a uniform input tensor contract across all \NumDatasets{} scenes. Users can freely
modify or bypass these configurations (classifying raw unreduced cubes, specifying custom spatial contexts,
or employing spatially disjoint partitions) directly through \texttt{config/config.yaml} without modifying source code.
The underlying implementation is detailed in \Cref{sec:datalayer}, and the configuration schema in \Cref{tab:protocol}.

One consequence is worth stating explicitly. The loader does not rebalance classes,
augment, denoise or drop noisy bands; it passes the label distribution through as
recorded, so the imbalance in \Cref{tab:datasets} reaches the model intact. Correction
strategies are themselves research questions, and a framework that applied one silently
would make its numbers incomparable with the literature it is meant to be read against.

Every scene is redistributed with attribution to its original provider, cited above, and
the consolidated collection is curated and hosted with attribution on Hugging Face;
individual scenes remain subject to the terms of their original
releases, which the collection card records per scene. No ground truth is modified, so a
result obtained here can be compared against one obtained from the original release.

\FloatBarrier
% =====================================================================
%  sec/4_pipeline.tex
% =====================================================================
\section{Hyperspectral-Image-Models: Software Architecture and Library Design}\label{sec:pipeline}

This section documents the software architecture and library design of Hyperspectral-Image-Models. Four
foundational design commitments shape the framework: the experiment is declared entirely in
one declarative configuration file, with nothing hard-coded or implicit in source code;
models are adapted only as far as a shared interface requires, with every adaptation
rigorously recorded alongside the code; complete run configurations and seeds are snapshotted
into every output directory for end-to-end provenance; and downstream artefacts (publication-grade
tables and figures) are generated programmatically rather than assembled manually.

\subsection{System architecture and library design}\label{sec:design}

\begin{figure}[t]
\centering
\includegraphics[width=\linewidth]{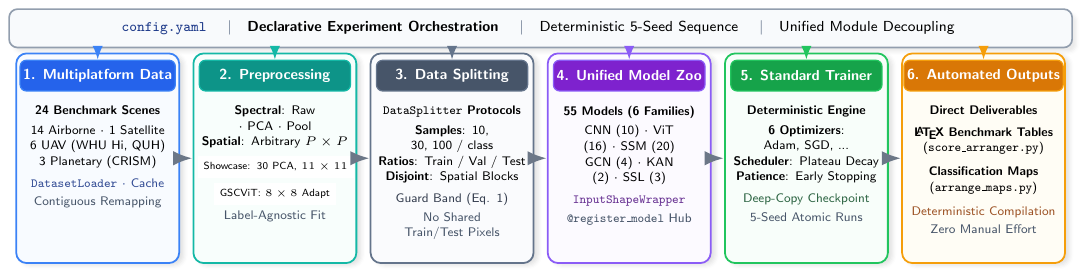}
\caption{The Hyperspectral-Image-Models execution path and modular framework architecture. A single declarative configuration file (\texttt{config.yaml}) controls dataset ingestion across \NumDatasets{} multiplatform scenes (14 Airborne, 1 Spaceborne, 6 UAV, 3 Planetary), spectral preprocessing (raw, PCA, or pooling), spatial patch extraction ($P \times P$), splitting strategies (balanced count, ratio, or disjoint blocks), and unified model execution across \NumCatalogModels{} catalog architectures spanning 6 paradigms. While the showcase evaluation in \Cref{sec:results} uses a standardized reference setting ($11 \times 11$, 30 PCA bands, 30/10 samples, 5 seeds) for fair comparison, the framework provides full user customizability across all pipeline stages, with automated emission of \LaTeX{} tables and thematic classification maps.}
\label{fig:pipeline}
\end{figure}

Analogous to how libraries like \texttt{timm} \citep{rw2019timm} and \texttt{segmentation\_models.pytorch}
\citep{Yakubovskiy:2019} unified 2D computer vision, Hyperspectral-Image-Models is built from the ground up as a
reusable, modular PyTorch library rather than an isolated benchmarking script. As shown in
\Cref{fig:pipeline}, the pipeline decouples data management, geometric tensor extraction,
architecture instantiation, training optimization, and artifact generation into cleanly partitioned modules.
A run proceeds as:
\begin{center}
\emph{Dataset} $\longrightarrow$ \emph{Preprocessing} $\longrightarrow$ \emph{Splitting} $\longrightarrow$ \emph{Model Registry} $\longrightarrow$ \emph{Trainer} $\longrightarrow$ \emph{Evaluation} $\longrightarrow$ \emph{Artifacts}
\end{center}

\begin{table}[t]
\centering
\caption{The Hyperspectral-Image-Models software architecture and modular capability matrix. The library decouples data ingestion, preprocessing, partitioning, model registration, training, and evaluation into interchangeable, configuration-driven components.}
\label{tab:capabilities}
\small
\setlength{\tabcolsep}{5pt}
\begin{tabularx}{\linewidth}{@{}l l >{\raggedright\arraybackslash}X@{}}
\toprule
Subsystem & Core Component & Modular Capabilities \\
\midrule
Dataset Layer & \texttt{DatasetLoader} & Unified retrieval and caching across 24 multi-platform scenes (Airborne, Spaceborne, UAV, Mars CRISM); automatic $[0,1]$ scaling, label remapping, background masking. \\
Preprocessing & \texttt{HyperspectralDataset} & Arbitrary $P \times P$ spatial patch extraction ($5\times5$ to $27\times27$); PCA dimensionality reduction to $C$ components, band pooling, explicit selection, or full raw spectra. \\
Split Engine & \texttt{data\_split.py} & Class-balanced random sampling (fixed counts or fractional percentage ratios) and spatially disjoint component-level partitioning with a guard band that removes evaluation patches overlapping any training patch. \\
Model Registry & \texttt{@register\_model} & Unified model ecosystem hosting 55 architectures across 6 paradigms (CNN, ViT, SSM/Mamba, GCN, KAN, MAE); auto-discovery; dynamic 4D/5D \texttt{InputShapeWrapper}. \\
Training & \texttt{trainer.py} & Shared loop with 6 optimizers (Adam, AdamW, SGD, RMSprop, Adagrad, Adadelta), plateau LR decay, validation early stopping, atomic checkpointing. \\
Evaluation & \texttt{metrics.py} & Strict held-out evaluation: OA, AA, Cohen's $\kappa$, full confusion matrices, per-class accuracies, and cross-seed aggregation ($\mu \pm \sigma$). \\
Complexity & \texttt{model\_info.py} & Parameter counts and multiply-accumulate operations (MACs/FLOPs) under uniform probe tensors $(1, 1, C, P, P)$. \\
Deliverables & Downstream Tools & Automated generation of paper-ready \LaTeX{} tables (\texttt{score\_arranger.py}) and arranged full-scene thematic maps (\texttt{arrange\_maps.py}). \\
Provenance & \texttt{config\_loader.py} & Global deterministic seed locking across CPU/CUDA/cuDNN; automatic per-run YAML configuration snapshots archived in every output folder. \\
\bottomrule
\end{tabularx}
\end{table}

\Cref{tab:capabilities} contrasts the architectural design of Hyperspectral-Image-Models against typical published
hyperspectral codebases. In standard practice, individual research papers release standalone scripts
tightly coupled to a single dataset, fixed patch size, and hardcoded spectral dimensionality.
In contrast, Hyperspectral-Image-Models exposes every stage through programmatic interfaces and YAML configurations,
enabling researchers to switch datasets, swap architectures, modify patch dimensions, or toggle
between spatial splitting paradigms with zero modifications to underlying model code.

\subsection{Data loading and preprocessing}\label{sec:datalayer}

\texttt{DatasetLoader} resolves a scene name against the registry in
\texttt{config/dataset.yaml}, downloads the collection entry if it is not cached, and
reads the MATLAB arrays, locating the data and ground-truth keys by inspection rather
than by a fixed naming convention, since the public releases do not agree on one. The
loaded cube is then matched in orientation and extent to its label map, and each band is
min-max scaled to $[0,1]$ over the scene-wide dynamic range,
$X' = (X - X_{\min})/(X_{\max} - X_{\min})$. The unlabelled background class is dropped
and the remaining labels are remapped to a contiguous range, so a scene whose released
ground truth uses non-contiguous indices needs no special handling downstream.

The preprocessing stage then handles the spectral axis. Three options are natively configurable:
principal component analysis (PCA) to a chosen number of components $C$, explicit band index
selection (applied after PCA when both are set), or retaining the full unreduced spectral profile.
A separate \texttt{maxpool} mode applies $k\times k$ max-pooling over the spatial axes of each patch
and leaves the spectral axis unchanged. PCA is seeded with the run seed, so for a given seed the
projection basis is identical across models; without this, competitors would evaluate against
divergent input representations. We explicitly note that, adhering to prevailing convention in the hyperspectral
benchmarking literature \citep{audebert2019deep}, both min-max normalization and PCA dimensionality
reduction are fitted globally over the complete spatial scene prior to pixel partitioning. This
guarantees a consistent, deterministic spectral coordinate basis across all candidate spatial windows
while remaining strictly label-agnostic, as neither transformation accesses ground-truth annotations.
\texttt{HyperspectralDataset} finally extracts spatial patches of arbitrary dimensions $P \times P$
(e.g., $7\times7, 11\times11, 15\times15, 27\times27$) from the valid spatial coordinate grid centred
on labelled pixels at the configured stride, without artificial border extrapolation, and emits
batches of shape $(B, 1, C, P, P)$ or $(B, C, P, P)$ per sample.

\subsection{Dataset splitting}\label{sec:splitting}

The framework natively supports arbitrary dataset partitioning through two distinct split
paradigms, governed entirely through \texttt{config/config.yaml}:

The \emph{standard} family performs class-balanced random sampling from the labelled pixel pool.
Users can declare splits either as continuous percentage ratios (\texttt{method: ratio}, such as
5\%/5\%/90\% or 10\%/10\%/80\% train/val/test splits) or as explicit integer sample budgets per
class (\texttt{method: samples}, such as 15, 30, or 100 samples per class), with the remaining
annotated pixels forming the test partition. Because sampling is performed per class, every category
is guaranteed representation during training regardless of scene-wide class imbalance.
Where available ground truth in a class is smaller than the requested budget, the loader stops with
an error rather than silently shrinking that class, and the budget is lowered in the configuration
(for example, on Indian Pines, where classes contain as few as 20 pixels, a 10/5 budget is set). \Cref{tab:datasets} illustrates the real-world sampling variation: a budget
of 30 training samples per class utilizes 12.93\% of the available ground truth in Botswana, but only
0.03\% in Houston 2018.

The \emph{disjoint} family separates training and evaluation in space rather than by
random draw. It targets evaluation under spatial shift and removes the window overlap
that inflates accuracy under the standard per-class split \citep{nalepa2019validating,liang2017sampling}. It is built in two stages. First, each class is
decomposed into its 4-connected components. A class with a single component is cut along
the longer axis of its bounding box into contiguous training, validation and test strips;
with two components the smaller one is assigned whole to the split whose target share it
best matches and the larger one is cut among the remaining splits; with three or more,
whole components are assigned greedily from the largest down, and the dominant training
component is re-cut if training exceeds its target by more than five percentage points.
Second, every patch is assigned to the split that contains its centre pixel. Target
ratios are set with \texttt{data\_split.split\_ratios} (50\%/30\%/20\% by default); because
whole components are assigned, the realised shares deviate from the targets and are
reported by the splitter for every run (\Cref{tab:disjointsplit}).

Centre-pixel assignment alone does not make the patches disjoint: a $P\times P$ test
patch centred within $P-1$ pixels (chessboard distance) of a training centre still
contains training pixels, and this happens along every region boundary, including
boundaries between the regions of different classes \citep{liang2017sampling}. The
framework therefore applies a \emph{guard band} by default
(\texttt{data\_split.guard\_band: true}). With $\mathcal{T}$ the set of training patch
centres, a validation or test patch centred at $\mathbf{c}$ is kept only if
\begin{equation}
\min_{\mathbf{t}\in\mathcal{T}} \lVert \mathbf{c}-\mathbf{t} \rVert_\infty > P-1,
\label{eq:guard}
\end{equation}
which is exactly the condition under which the two $P\times P$ windows share no pixel.
Every other validation or test patch is removed, so no pixel observed during training is
observed again during evaluation. The condition is evaluated with one chessboard distance
transform of the training-centre mask, so its cost is linear in the scene size. The number of patches
removed per split is logged.

Two further behaviours are made explicit rather than hidden.
If a class retains fewer than five training or test samples after assignment, samples are
moved across regions to reach that minimum, and each move (class, count, source and
destination split) is printed so it can be reported. And because the region masks are a
deterministic function of the ground truth, repeated seeds share one geometric partition;
under the disjoint protocol the reported standard deviation therefore reflects
initialisation and optimisation noise, not split variation.

A sample-budget variant
(\texttt{method: samples} with \texttt{disjoint: true}) draws the per-class budget from a
disjoint training region and evaluates on the opposite region, leaving unpicked
training-region samples out of evaluation altogether; in this variant the validation
samples are drawn from the evaluation region, so early stopping observes the test
distribution, although never a test pixel. Because the guard band acts after the
validation samples are drawn, the realised validation budget can fall below the requested
one, and a class can lose all of its validation or test patches; the splitter reports both
cases. Finally, \texttt{tools/disjoint\_audit.py} computes, from the ground truth alone and
without training, how many test windows overlap a training window under any protocol, the
realised disjoint shares, the guard-band removals and a map of each split.

The showcase benchmark in \Cref{sec:results} uses the standard per-class protocol (30
training and 10 validation samples) for comparability with the bulk of published
results; the disjoint protocol is a one-line switch (\texttt{data\_split.disjoint: true})
intended for evaluation under spatial shift.

\subsection{Model interface and extensible ecosystem}\label{sec:registry}

A core contribution of Hyperspectral-Image-Models is establishing a unified model ecosystem for hyperspectral
classification. In mainstream computer vision, models share standard $(B, 3, H, W)$ image tensors.
In hyperspectral imaging, however, architectures historically diverge between two mutually incompatible
tensor conventions:
\begin{itemize}
    \item \textbf{5D Tensor Convention} $(B, 1, C, H, W)$: Utilized by 3D convolutional networks and
    spectral-spatial state-space models that treat spectral bands as a third spatial depth dimension.
    \item \textbf{4D Tensor Convention} $(B, C, H, W)$: Utilized by 2D CNNs, Vision Transformers, and
    Mamba architectures that treat spectral bands as feature channels.
\end{itemize}

Hyperspectral-Image-Models harmonizes these conventions through a strict factory contract and an automated adapter.
Every registered architecture factory must implement a uniform constructor signature accepting
\texttt{num\_classes}, \texttt{bands}, and \texttt{patch\_size}:
\begin{lstlisting}[style=shellstyle,caption={Model registration and constructor signature.},label={lst:register}]
@register_model('ModelName', expects_4d=True, hidden_dim=64)
def factory(num_classes, bands, patch_size=11, hidden_dim=64, **kwargs):
    return ModelName(num_classes=num_classes, bands=bands, patch_size=patch_size, ...)
\end{lstlisting}

Models adhering to the 4D convention declare \texttt{expects\_4d=True} in their registration decorator.
The runner automatically wraps the model in \texttt{InputShapeWrapper}, which inspects incoming batches
at runtime: if a 5D tensor with a singleton depth axis $(B, 1, C, H, W)$ is supplied by the data loader,
it squeezes dimension 1 to produce $(B, C, H, W)$, and otherwise passes the tensor transparently.
Furthermore, all models emit unnormalized raw logits $(B, \text{num\_classes})$ rather than softmax
probabilities, guaranteeing that loss computation, metric evaluation, and gradient updates remain
strictly uniform.

\paragraph{Zero-touch extensibility: adding a new model.}
Contributing a new architecture to the Hyperspectral-Image-Models ecosystem requires no modification to the core
training engine, data loaders, or CLI tools. A researcher follows a 4-step workflow:
\begin{enumerate}
    \item Place the PyTorch model file under \texttt{models/yYYYY/} according to its publication year.
    \item Decorate the factory function with \texttt{@register\_model('ModelName', expects\_4d=...)}.
    \item Add the bibliographic metadata (title, DOI, authors, venue) to \texttt{MODEL\_CATALOG} in \texttt{models/registry.py}.
    \item Document any adaptations from the original paper's release in the module header docstring.
\end{enumerate}
Once registered, the model is immediately available for single runs, sweeps, and benchmarking via
the CLI (\Cref{lst:cli}).

\subsection{Training configuration}\label{sec:training}

The trainer is shared by every model. It exposes six optimisers (Adam, AdamW, SGD,
RMSprop, Adagrad and Adadelta), cross-entropy loss, plateau-based learning-rate decay,
early stopping on validation accuracy, and checkpointing of the best and final states.
Epoch budget, batch size, learning rate, optimiser and early-stopping patience are
configuration keys; the plateau schedule (factor 0.5, patience 5, on validation loss) is fixed in
\texttt{utils/trainer.py}. The values used for the benchmark are stated in
\Cref{tab:protocol}; the point is not that these are optimal for any individual
architecture but that they are the same for all of them, which is the condition under
which an accuracy difference can be attributed to the architecture.

\subsection{Evaluation protocol and metrics}\label{sec:metrics}

On completion of training, the retained best checkpoint is evaluated on the test
partition, which is disjoint from both the training and validation partitions by
construction. Let $M \in \mathbb{N}^{K \times K}$ be the confusion matrix over $K$
classes, with $M_{ij}$ the number of test pixels of true class $i$ predicted as class
$j$, and $N = \sum_{ij} M_{ij}$. The framework reports three scalar metrics:

\begin{equation}
\mathrm{OA} = \frac{1}{N}\sum_{i=1}^{K} M_{ii},
\qquad
\mathrm{AA} = \frac{1}{K}\sum_{i=1}^{K} \frac{M_{ii}}{\sum_{j} M_{ij}},
\qquad
\kappa = \frac{p_o - p_e}{1 - p_e},
\label{eq:metrics}
\end{equation}

where $p_o = \mathrm{OA}$ and
$p_e = N^{-2}\sum_{i} \left(\sum_{j} M_{ij}\right)\left(\sum_{j} M_{ji}\right)$ is the
agreement expected by chance. The three are not redundant. Overall accuracy is dominated
by the largest classes; average accuracy weights every class equally and therefore
exposes minority-class failure; and $\kappa$ discounts the agreement a trivial classifier
would obtain from the class priors alone. Given the imbalance ratios in
\Cref{tab:datasets}, which reach $1{,}524\times$ on Houston 2018, reporting only one of
the three would be misleading. Per-class accuracies are written alongside them for every run.

Model complexity is measured separately from training. \texttt{model\_info.py} profiles
trainable parameters and multiply-accumulate operations for every registered architecture
under one fixed probe input, so the numbers are comparable across models. They are
structural properties measured under identical conditions, not performance measurements.

\subsection{Reproducibility and configuration}\label{sec:repro}

Repetition is treated as part of an experiment rather than as an afterthought. Each model
and scene pair is run once per entry in the seed list, and a seed is applied globally
before the split is drawn, so Python's \texttt{random}, NumPy and PyTorch on both CPU and
CUDA are seeded together and cuDNN runs in deterministic mode with the autotuner
disabled. Two consequences matter for comparability: repeated runs of the same model are
reproducible, and, more importantly, different models sharing a seed list see identical
splits and identical initialisation conditions. Results are reported as a mean with a
sample standard deviation over the repeats rather than as a best run, following standard
reproducibility practice \citep{pineau2021repro}.

\Cref{tab:protocol} compiles the standardized unified protocol adopted specifically to
showcase the benchmark results across all \NumCatalogModels{} catalog architectures and \NumDatasets{} scenes in the
Hyperspectral-Image-Models framework under identical, commensurable conditions. The selected values, an $11\times11$
spatial window, 30 retained principal components, and a reference budget of 30 labelled training and 10 validation
samples per class (with an adaptive reduction to 10 training and 5 validation samples per class on Indian Pines
to accommodate its smallest categories), sit at the centre of common literature practice, providing a grounded reference
point across paradigms. Crucially, nothing in the Hyperspectral-Image-Models framework is constrained or
hardcoded to this specific configuration: every entry in \Cref{tab:protocol} corresponds to
a first-class configuration key in \texttt{config/config.yaml}. Researchers can freely specify
custom patch dimensions ($P \times P$), arbitrary spectral dimensionality reductions or raw spectra,
alternative sampling budgets or percentage ratios, spatially disjoint geographic partitions,
and customized optimization routines. What is essential for the benchmark numbers reported here
is that they were held strictly constant across all models and scenes, and that every parameter
is archived with end-to-end experimental provenance.

%% Values transcribed from config/config.yaml, utils/trainer.py and
%% utils/experiment_runner.py in the released repository.
\begin{table}[t]
\centering
\caption{The standardized unified protocol used to showcase benchmark results across all \NumCatalogModels{} models and \NumDatasets{} scenes in the Hyperspectral-Image-Models framework. Every value is read from \texttt{config/config.yaml} unless the source column says otherwise, and the configuration file is copied into each run directory for complete provenance. Holding these parameters constant enables fair, commensurable cross-model comparison; researchers can freely adapt any setting (including patch size, spectral band reduction, sampling budgets, spatial disjoint partitioning, and optimizer schedules) to suit their own model compositions and experimental protocols.}
\label{tab:protocol}
\small
\setlength{\tabcolsep}{4pt}
\begin{tabularx}{\linewidth}{@{}l >{\raggedright\arraybackslash}X >{\raggedright\arraybackslash}p{0.40\linewidth}@{}}
\toprule
Setting & Value & Source \\
\midrule
\multicolumn{3}{l}{\textit{Input}} \\[1pt]
\quad Patch size            & $11\times11$              & \texttt{dataset.patch\_size} \\
\quad Patch stride          & 1                         & \texttt{dataset.stride} \\
\quad Dimensionality reduction & PCA                    & \texttt{preprocessing.dim\_reduction\_method} \\
\quad Retained components   & 30                        & \texttt{preprocessing.num\_pca\_bands} \\
\quad Tensor layout         & $(1, \text{bands}, H, W)$ & \texttt{preprocessing.use\_channel\_dim} \\
\quad Band normalisation    & per-band min--max to $[0,1]$ & \texttt{utils/data\_loader.py} \\
\addlinespace
\multicolumn{3}{l}{\textit{Split}} \\[1pt]
\quad Method                & fixed count per class     & \texttt{data\_split.method} \\
\quad Train / validation    & 30 / 10 per class (Indian Pines: 10 / 5); remainder test & \texttt{data\_split.split\_samples} \\
\quad Spatially disjoint    & available, off by default & \texttt{data\_split.disjoint} \\
\quad Guard band            & on whenever disjoint is enabled & \texttt{data\_split.guard\_band} \\
\quad Base seed             & 0                         & \texttt{data\_split.random\_state} \\
\quad Per-run seeds         & explicit list, else base $+\,(n-1)$ & \texttt{utils/experiment\_runner.py} \\
\quad Repeats               & 5 runs per model per scene & \texttt{training.num\_runs} \\
\addlinespace
\multicolumn{3}{l}{\textit{Optimisation}} \\[1pt]
\quad Epochs                & 100                       & \texttt{training.num\_epochs} \\
\quad Batch size            & 64                        & \texttt{training.batch\_size} \\
\quad Optimiser             & Adam                      & \texttt{training.optimizer} \\
\quad Learning rate         & $1\times10^{-3}$          & \texttt{training.learning\_rate} \\
\quad Schedule              & plateau decay, factor 0.5, patience 5 & \texttt{utils/trainer.py} \\
\quad Early stopping        & patience 10 on validation accuracy & \texttt{training.patience} \\
\quad Loss                  & cross-entropy             & \texttt{utils/trainer.py} \\
\addlinespace
\multicolumn{3}{l}{\textit{Determinism}} \\[1pt]
\quad Seeded                & Python, NumPy, Torch CPU and CUDA & \texttt{set\_global\_seed} \\
\quad cuDNN                 & deterministic on, autotuner off & \texttt{set\_global\_seed} \\
\quad PCA                   & seeded with the run seed  & \texttt{utils/experiment.py} \\
\addlinespace
\multicolumn{3}{l}{\textit{Complexity probe}} \\[1pt]
\quad Probe input           & $(1,1,30,11,11)$, 16 classes & \texttt{model\_info.py} \\
\bottomrule
\end{tabularx}
\end{table}

\subsection{Models implemented in Hyperspectral-Image-Models}\label{sec:zoo}

\Cref{tab:families} summarises the implemented collection by paradigm; the complete
catalogue, with venue, identifier, parameter count, operation count and admissible patch
size for every entry, is \Cref{tab:models} in \Cref{app:models}. Architectural credit for
every entry belongs to its original authors, and the framework claims the integration
rather than the architectures.

\begin{table}[t]
\centering
\caption{The implemented collection by paradigm. Parameter and operation counts are measured under one fixed probe input and span three and four orders of magnitude respectively; every entry takes a hyperspectral patch as its only input and is trained under the single protocol of \Cref{tab:protocol}. \Cref{tab:models} in \Cref{app:models} gives the per-model breakdown with venues, identifiers and costs. The family rows and cost spans are aggregated automatically by the Hyperspectral-Image-Models framework from the same run records and structural profile used to emit \Cref{tab:perscene}.}
\label{tab:families}
\small
\setlength{\tabcolsep}{4pt}
\begin{tabular}{@{}lrcrrp{0.35\linewidth}@{}}
\toprule
Paradigm & Impl. & Years & Par. (M) & MACs (M) & Core mechanism \\
\midrule
CNN & 10 & 2017--2026 & 0.012--1.34 & 1.7--102 & local weight sharing, 2D/3D spectral-spatial convolution \\
Transformer & 16 & 2021--2026 & 0.003--4.44 & 1.8--615 & multi-head self-attention over spectral or patch tokens \\
Mamba / SSM & 20 & 2024--2026 & 0.052--4.25 & 0.2--345 & input-dependent selective state-space scan, linear in length \\
Graph / GCN & 4 & 2024--2026 & 0.022--0.33 & 1.9--28 & message passing over a superpixel adjacency graph \\
KAN & 2 & 2024--2024 & 0.344--0.43 & 1.2--11 & learnable univariate B-spline functions on network edges \\
Self-supervised & 3 & 2024--2024 & 0.513--34.23 & 16.3--4195 & masked spectral-spatial reconstruction, then fine-tuning \\
\midrule
\textbf{Total} & \textbf{55} & \YearSpan{} & & & \\
\bottomrule
\end{tabular}
\end{table}

The composition by year reflects the first public release or preprint appearance of each architecture
(with formal archival journal or conference publication details following the cited references):
convolutional designs dominate through 2020, transformers from 2021 to 2023, and state-space models from 2024,
with graph, Kolmogorov-Arnold and self-supervised entries appearing alongside rather than
displacing them. Cost spans a wider range than the paradigm labels suggest, from
0.003\,M parameters to 34.2\,M and from under 1\,MMAC to 4.2\,GMAC, and that range cuts
across families rather than separating them. Because these diverse paradigms are present in one
environment, questions that previously required reimplementation, such as how a
Mars-specific graph model behaves on terrestrial scenes, become configuration changes.

\subsection{Integration: adapting published code to a shared interface}\label{sec:integration}

Every codebase gathered into the framework is internally consistent and locally
reasonable: each was written against one loader, one patch size, one machine and one
dependency set, and inside that setting its choices are the sensible ones. The friction
below appears only when \NumCatalogModels{} such codebases must coexist behind one
interface and one seeded protocol. It is a property of independently developed research
code rather than a fault of its authors, and it is reported because it is the part of a
benchmark that decides whether its numbers can be trusted. The full record is maintained
directly in the repository documentation and model module docstrings.

\paragraph{Hard-coded scene assumptions.}
A model written for one scene has no reason to parameterise what that scene fixes.
\texttt{HSIC\_SClusterFormer} carried a spectral width of 30 as a literal in several
modules, so the port threads an explicit \texttt{pca\_components} argument through the
affected constructors, and \texttt{HyperKAN}'s 2D convolution took an input width that
followed from the band count, now computed at construction time. \texttt{HSIConvKAN} is
the sharpest case: its reference notebook pairs a fixed \texttt{MaxPool2d} of kernel and
stride 3 with a fixed classifier head, a combination consistent only at the patch size
that notebook uses, since pooling a height-one feature map with a kernel of three fails
outright; an \texttt{AdaptiveMaxPool2d} reaches the same head width from any input size.

\paragraph{Structural input constraints.}
Some constraints are not incidental. \texttt{GSCViT} asserts that spatial dimensions
are divisible by every entry of its group size, $[4,4,4]$ in every published
configuration, and since spatial dimensions are preserved through all stages, this
condition applies at every depth. The requirement is divisibility rather than one
admissible resolution; thus 8, 12, 16, 20, and 24 satisfy it, whereas the protocol's default 11 does not.
Rather than forcing a non-divisible patch into one group (which would collapse the multi-level
grouped attention hierarchy that defines the architecture), the framework evaluates \texttt{GSCViT}
with an $8\times8$ spatial window, the nearest dyadic size satisfying its grouping constraints.
This transparent protocol adaptation enables \texttt{GSCViT} to complete all \NumDatasets{} benchmark scenes
while preserving its intended architectural inductive biases, exemplifying the framework's principle
of making model-specific adaptations explicit and reproducible rather than silently altering network semantics.

\paragraph{Dependency weight.}
The state-space family carries the heaviest installation cost. \texttt{mamba\_ssm} and
\texttt{causal-conv1d} compile CUDA kernels, must be installed without build isolation,
and have no CPU fallback \citep{gu2024mamba}. Two models need kernels that are not in the
published packages at all: \texttt{HyperMamba} calls VMamba's compiled selective scan
\citep{liu2024vmamba} and \texttt{EMamba} calls one built in its own repository. Both are
replaced with \texttt{selective\_scan\_ref}, the algebraically equivalent reference
implementation shipped with \texttt{mamba\_ssm}. \texttt{HyperMamba} additionally used
\texttt{mmcv.ops.DeformRoIPool}; called with no offset tensor, as the original calls it,
that operator reduces to plain box RoI pooling, so the port uses
\texttt{torchvision.ops.roi\_align} over the same boxes and drops the dependency. The
framework's standing policy is to skip a model with a warning when an optional dependency
is unavailable, so an environment without compiled kernels loses those entries from the
listing rather than losing the run.

\paragraph{Device, shape and framework assumptions.}
Assumptions about where a tensor lives, what shape it has, or which framework surrounds
it are invisible until the code runs somewhere it was not written. \texttt{MambaLG}
called \texttt{.cuda()} inside \texttt{forward} and \texttt{pResNet} allocated its
channel-matching zero-pad the same way, so both now take their placement from the runner;
\texttt{pResNet}'s shortcut pool also floored the spatial size where the main path ceiled
it, failing at any odd patch size, which ceiling mode and an adaptive final pool resolve.
\texttt{ConvVitMamba} came from TensorFlow and Keras and was ported layer for layer,
except at attention, where Keras sets the query, key and value width per head
independently of the embedding width, so the port carries a module that reproduces the
original widths and emits raw logits instead of the original softmax. \texttt{EMamba}
required a reduction of a different kind: it is a hyperspectral and LiDAR fusion model
and the protocol is single-modality, so the port keeps only its hyperspectral stream and
drops the LiDAR branch with both fusion blocks, adding nothing in their place. The
complexity probe needed two fixes of its own, since \texttt{InputShapeWrapper} defeats a
profiler's attribute check and lazily created submodules are invisible when the module
tree is walked; profiling the raw model after a warm-up pass addresses both.

\paragraph{Fidelity policy.}
None of the above is useful unless it is recorded. \texttt{docs/CONTRIBUTING.md} makes a
deviation note in the module docstring a condition of adding a model, so each adaptation
sits next to the code it changes and can be read against the original release. The same
discipline surfaced the one latent bug the process found: the single-group branch of
\texttt{GSCViT}'s grouped attention names the wrong tensor axis and would raise if it
were reached, which it never is in the original because the divisibility assertion above
it excludes every input that would take that path. That is the character of what a shared
interface exposes. It does not find errors so much as it moves code into configurations
its authors had no reason to try, and the value lies in the record of what changed, which
is what makes a result obtained here reproducible elsewhere.

\subsection{Using the framework}\label{sec:usage}

Three keys decide what runs: \texttt{dataset.names} lists the scenes,
\texttt{model.name} lists the architectures, and \texttt{data\_split.seeds} lists the
seeds, one per repeat, and the experiment is their cross product. Two switches turn the
lists into sweeps, \texttt{run\_all\_datasets} and \texttt{run\_all\_models}, and
\texttt{model.exclude} covers the common case of running everything except a few. The
full schema, with the values that define the protocol, is given in \Cref{app:config}.

One entry point then covers ordinary use. The listing commands print what the registry and
the dataset catalogue currently contain; a bare invocation trains everything named in the
configuration, downloading any missing scene on first use; and the remaining commands
operate on completed runs. For sweeps across several configuration variants,
\texttt{run\_all\_experiments.py} drives the same entry point in a loop.

\begin{lstlisting}[style=shellstyle,caption={The commands that cover ordinary use.},label={lst:cli}]
python main.py --list-models     # what is registered
python main.py --list-datasets   # what scenes are available
python main.py                   # train everything in the config
python main.py --arrange-scores  # accuracy tables from completed runs
python main.py --arrange-only    # arranged classification-map figure
python model_info.py --latex     # parameter and operation-count table
\end{lstlisting}

Outputs go to \texttt{\{results.directory\}/\{dataset\}/\{model\}/run\_N/}, each holding
the best and final checkpoints, that run's configuration snapshot, the epoch-level
training log and, when requested, a classification map, with a
\texttt{results\_summary.csv} aggregating a model's runs. Because a full sweep produces
many checkpoints, only the best and worst run of a model retain their weights; every
snapshot, log and summary is kept, and the snapshot is what makes an output traceable.

As formalized in \Cref{sec:registry}, adding an architecture to the library requires only
four steps: placing the file in its publication year folder, registering it with
\texttt{@register\_model}, adding bibliographic metadata to \texttt{MODEL\_CATALOG}, and
recording any adaptations in the module docstring. \Cref{app:adding} works through these
steps end to end.

\FloatBarrier
% =====================================================================
%  sec/5_results.tex
% =====================================================================
\section{Experimental Results and Visual Analysis}\label{sec:results}

This section presents the empirical findings produced when the framework is evaluated at scale
across all \NumCatalogModels{} architectures and \NumDatasets{} scenes. The primary objectives are
to validate the framework's end-to-end execution, to document the automated benchmarking artefacts
it generates, and to establish a comprehensive, standardized reference showcase benchmark that future hyperspectral
investigations can build upon. To showcase these benchmark results under controlled, reproducible
conditions, all experiments reported here adhere strictly to the unified reference protocol of
\Cref{tab:protocol}, ensuring that every model evaluates on identical pixel partitions and operates
under uniform optimization constraints. We reiterate that while this showcase evaluation standardizes
on an $11\times11$ spatial window, 30 PCA bands, and a reference budget of 30 samples per class
(10 on Indian Pines), researchers deploying Hyperspectral-Image-Models can readily configure alternative patch
dimensions, raw spectral inputs, and custom split strategies via \texttt{config/config.yaml}.

\subsection{Experimental setup}\label{sec:setup}

All showcase benchmark runs strictly follow the reference configuration in \Cref{tab:protocol},
guaranteeing that every architecture uses the same labelled pixel partitions across the shared seed sequence,
with model-specific input adaptations applied where required by architectural constraints. Of the \NumCatalogModels{} catalogued architectures, \NumEvaluatedModels{}
have completed the full sweep across all \NumDatasets{} scenes, yielding \NumTotalEvaluations{}
model--scene combinations and \NumTotalRuns{} individual training runs. To prevent reporting artifacts from lucky
initializations, all reported performance metrics represent the mean over the \NumRepeats{} independent
seeded runs accompanied by the sample standard deviation ($\mu \pm \sigma$), rather than an isolated best run.

\subsection{Quantitative results}\label{sec:landscape}

The full evaluation matrix, \NumEvaluatedModels{} architectures against \NumDatasets{} scenes, shows
two structures at very different magnitudes.

The dominant one is vertical: scene difficulty explains far more of the variation than
architecture does. Botswana (mean OA \EasiestMeanOA\%), Pavia Centre (96.31\%), WHU Hi LongKou
(95.13\%), Chikusei (95.03\%) and Holden (94.82\%) have large, spectrally distinct
parcels, and nearly every architecture exceeds 90\% on them, while Houston 2018
(\HardestMeanOA\%), Berlin (64.75\%), Loukia (66.77\%) and Indian Pines (71.66\%) degrade
every paradigm at once. The spread between the easiest and hardest scene, \SceneSpread{}
accuracy points, is more than twice the \FamilySpread{}-point spread between the best and
worst family mean. This is the most consequential observation in the benchmark for how hyperspectral
results should be read: a method evaluated only on the classical scenes has been measured
on the easy end of the range.

The secondary structure is horizontal. Some architectures hold their accuracy across both
ends of the difficulty range, while others fall below 40\% on the hard scenes while
performing normally on the easy ones. \Cref{tab:dataset_difficulty} in
\Cref{app:difficulty} gives the difficulty spectrum for all \NumDatasets{} scenes, with
the observed accuracy range on each and the architecture attaining the highest value
there.

Across the \NumDatasets{} scenes the highest observed accuracy is distributed over four
paradigms: state-space models lead on 8 scenes, CNNs on 7, transformers on 6 and graph
networks on 3. No paradigm holds a universal advantage, and the one that leads tracks
scene geometry rather than publication year: graph models where parcels are large and
contiguous, transformers and compact CNNs where they are fine and dense, and the two
hardest urban scenes to a 2020 convolutional design and a 2024 hybrid transformer.

\subsection{Comparison across models}\label{sec:families}

Aggregating the benchmark performance by architectural paradigm reveals that the result is best
read as a statement about dispersion rather than about ranking. Convolutional networks (\CNNMeanOA\% mean, $\sigma = \CNNStdOA\%$) and graph networks
(\GraphMeanOA\%, $\sigma = \GraphStdOA\%$) have the highest family means and the tightest
distributions in the collection. Transformers reach \TransformerMeanOA\% with a much wider spread
($\sigma = \TransformerStdOA\%$), largely because of one weak member (DBCTNet, 35.12\%), while
their best entries are close to the top of the benchmark. State-space models reach \MambaMeanOA\%
($\sigma = \MambaStdOA\%$): the strongest entries, R2Mamba at 90.31\%, S2Mamba at 89.91\% and
IGroupSS-Mamba at 89.87\%, match the best models of any family, but the family as a whole is more
sensitive to scene geometry. The two Kolmogorov-Arnold entries sit at
\KANMeanOA\%, within the convolutional range.

The graph result deserves a closer look, because all four evaluated graph models land between 82.38\% and 88.95\% (GTCFN at 88.95\%, MS2GCAN at 88.83\%, MCTGCL at 86.74\% and GraphGST at 82.38\%), with MS2GCAN among the most consistent architectures in the entire \NumDatasets{}-scene benchmark ($\sigma = 9.62$ across scenes, reaching 99.84\% on Botswana and 99.55\% on Pavia Centre). In contrast, the self-supervised family mean (\SelfSupMeanOA\%, $\sigma = \SelfSupStdOA\%$) should be read with one caveat: in this showcase the three self-supervised backbones are trained from scratch with the same supervised loss as every other model, without their self-supervised pretraining stage or pretrained weights. Under that setting LFSMIM (84.29\%) and HSIMAE (82.95\%) perform robustly, while the foundation-scale HSIC\_FM averages 48.82\% in this small-sample (30 samples/class) regime.

\begin{figure}[t]
\centering
\includegraphics[width=\linewidth]{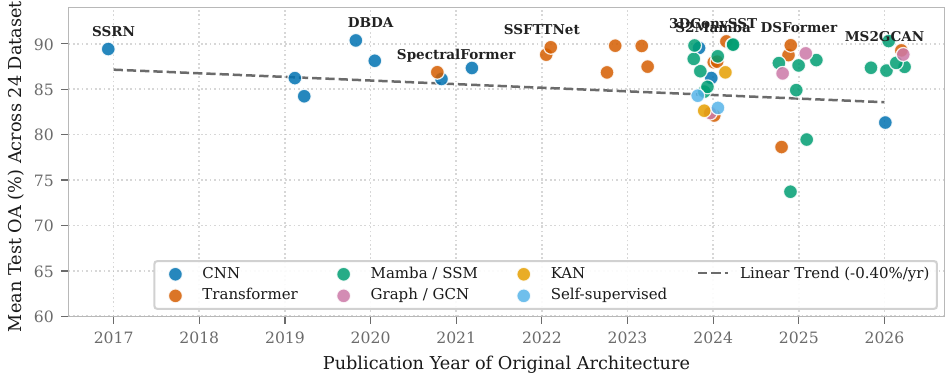}
\caption{Mean overall accuracy across \NumDatasets{} scenes against publication year,
coloured by paradigm, with a linear trend. The trend indicates the magnitude of the
year effect relative to within-year variation; it is not intended for extrapolation.}
\label{fig:generations}
\end{figure}

Plotting mean accuracy against publication year (\Cref{fig:generations}) gives a fitted
slope of $\YearSlope$ percentage points per year across the decade. Rather than indicating
architectural regression, this negative trend reflects the shifting composition of the field:
early designs (2017--2020) were refined, supervised CNNs tightly optimized for small-sample
classification on classical scenes, whereas recent cohorts (2024--2026) include ambitious,
exploratory paradigms such as foundation-scale models trained here without their pretraining (HSIC\_FM at 48.82\%) and
specialized state-space or graph variants that exhibit much wider performance dispersion.
Crucially, this trend should be read against the spread within a single year: the twenty-two
architectures published in 2024 alone range from 35.12\% (DBCTNet) to 90.27\% (3DConvSST), a spread of
more than 55 points. Convolutional designs from 2017 to 2020 span 84.23\% to 90.37\%, and the top of
that range, DBDA (2020), is the highest mean in the entire benchmark. Transformers introduced
between 2021 and 2023 span 86.85\% to 89.79\%, the tightest band of any generation. The 2024
to 2026 cohort is by far the most internally varied, containing both the strongest state-space
entries and the weakest results in the collection.

Mean accuracy alone does not describe an architecture's usefulness, since a model that is
excellent on half the scenes and poor on the rest may share a mean with one that is
uniformly good. Cross-scene standard deviation separates the two. Seventeen architectures combine a mean above
88\% with a standard deviation below 10.5 points, and twelve of those hold $\sigma$ below
10: DBDA, 3DConvSST, R2Mamba, S2Mamba, IGroupSS-Mamba, DSFormer, MambaLG, MorphFormer,
SSFTTNet, DKDMN, MS2GCAN and CTMixer. These are the architectures whose behaviour on an
unseen scene is most predictable. At the other extreme, WaveMamba
($\sigma = 20.08$) ranges from 85.27\% on Holden to 16.18\% on MUUFL, and HSIC\_FM
($\sigma = 19.89$), MHSSMamba ($\sigma = 19.23$) and DBCTNet ($\sigma = 18.08$) are similarly scene dependent.
Dispersion is not a property of a family: the most and the least consistent state-space
models sit at opposite ends of this range.

\subsection{Detailed results on four representative scenes}\label{sec:perscene}

Aggregates over \NumDatasets{} scenes are useful for characterising the collection but cannot be
compared against published numbers, which are almost always reported per scene. We
therefore give the complete per-model results, in all three metrics, for four scenes
chosen to span the difficulty and platform range of the collection.
\Cref{tab:perscene} covers Indian Pines and Pavia University, the two most frequently
reported benchmarks in the literature, WHU Hi LongKou, a UAV agricultural mosaic, and
Nili Fossae, a Mars CRISM scene.

\begin{table}[p]
\centering
\caption{Per-model overall and average accuracy on four representative scenes under the protocol of \Cref{tab:protocol}: two classical airborne benchmarks, one UAV agricultural mosaic and one Mars CRISM scene. Each entry is the mean over \NumRepeats{} seeded runs with its sample standard deviation, and the best value in each column is in bold. Cohen's $\kappa$ is omitted for width and because it correlates with overall accuracy at $r = \OAKappaCorr$ across the benchmark; all three metrics are recorded for every run. Models are ordered within each family by their summed accuracy over the four scenes. The table is emitted automatically by the Hyperspectral-Image-Models framework from the completed runs of the same repository released with this paper.}
\label{tab:perscene}
\scriptsize
\setlength{\tabcolsep}{3pt}
\begin{tabular}{@{}lrrrrrrrr@{}}
\toprule
& \multicolumn{2}{c}{Indian Pines} & \multicolumn{2}{c}{Pavia University} & \multicolumn{2}{c}{WHU-Hi-LongKou} & \multicolumn{2}{c}{Nili Fossae} \\
\cmidrule(lr){2-3} \cmidrule(lr){4-5} \cmidrule(lr){6-7} \cmidrule(lr){8-9}
Model & OA & AA & OA & AA & OA & AA & OA & AA \\
\midrule
\multicolumn{9}{@{}l}{\textit{CNN}} \\[1pt]
\quad \texttt{SSRN} & 84.71\,\tiny{$\pm$\,3.9} & \textbf{91.99\,\tiny{$\pm$\,1.4}} & \textbf{97.81\,\tiny{$\pm$\,0.5}} & \textbf{98.09\,\tiny{$\pm$\,0.4}} & 98.36\,\tiny{$\pm$\,0.3} & 97.44\,\tiny{$\pm$\,0.4} & 96.69\,\tiny{$\pm$\,1.0} & 95.29\,\tiny{$\pm$\,1.8} \\
\quad \texttt{DBDA} & 80.73\,\tiny{$\pm$\,3.9} & 89.34\,\tiny{$\pm$\,1.7} & 97.46\,\tiny{$\pm$\,0.3} & 97.29\,\tiny{$\pm$\,0.5} & 98.05\,\tiny{$\pm$\,0.2} & 96.78\,\tiny{$\pm$\,0.2} & 96.45\,\tiny{$\pm$\,0.5} & 95.76\,\tiny{$\pm$\,0.7} \\
\quad \texttt{DKDMN} & 84.82\,\tiny{$\pm$\,1.8} & 91.64\,\tiny{$\pm$\,1.1} & 95.44\,\tiny{$\pm$\,1.2} & 95.82\,\tiny{$\pm$\,1.7} & 97.64\,\tiny{$\pm$\,0.5} & 95.90\,\tiny{$\pm$\,0.8} & 93.73\,\tiny{$\pm$\,1.4} & 94.46\,\tiny{$\pm$\,0.9} \\
\quad \texttt{ENL\_FCN} & 80.73\,\tiny{$\pm$\,2.2} & 90.37\,\tiny{$\pm$\,1.3} & 94.43\,\tiny{$\pm$\,1.7} & 95.32\,\tiny{$\pm$\,0.6} & 96.93\,\tiny{$\pm$\,0.6} & 95.58\,\tiny{$\pm$\,1.4} & 92.22\,\tiny{$\pm$\,3.6} & 93.26\,\tiny{$\pm$\,2.4} \\
\quad \texttt{FETNet} & 80.84\,\tiny{$\pm$\,1.3} & 89.29\,\tiny{$\pm$\,0.6} & 87.36\,\tiny{$\pm$\,3.9} & 90.30\,\tiny{$\pm$\,1.9} & 97.31\,\tiny{$\pm$\,0.9} & 97.14\,\tiny{$\pm$\,0.4} & 96.19\,\tiny{$\pm$\,1.4} & 94.03\,\tiny{$\pm$\,2.2} \\
\quad \texttt{SSTN} & 73.10\,\tiny{$\pm$\,4.6} & 84.54\,\tiny{$\pm$\,4.9} & 94.18\,\tiny{$\pm$\,1.3} & 94.79\,\tiny{$\pm$\,0.7} & 97.20\,\tiny{$\pm$\,0.3} & 95.33\,\tiny{$\pm$\,0.5} & 96.11\,\tiny{$\pm$\,0.4} & 95.21\,\tiny{$\pm$\,0.7} \\
\quad \texttt{pResNet} & 66.82\,\tiny{$\pm$\,8.4} & 79.77\,\tiny{$\pm$\,4.9} & 94.10\,\tiny{$\pm$\,3.8} & 95.20\,\tiny{$\pm$\,0.6} & 97.73\,\tiny{$\pm$\,1.2} & 97.15\,\tiny{$\pm$\,1.6} & 97.06\,\tiny{$\pm$\,0.6} & 95.90\,\tiny{$\pm$\,1.1} \\
\quad \texttt{S3ANet} & 69.59\,\tiny{$\pm$\,2.1} & 80.61\,\tiny{$\pm$\,1.9} & 92.23\,\tiny{$\pm$\,1.5} & 92.83\,\tiny{$\pm$\,0.5} & 96.82\,\tiny{$\pm$\,0.5} & 96.33\,\tiny{$\pm$\,0.3} & 96.94\,\tiny{$\pm$\,1.3} & 97.01\,\tiny{$\pm$\,0.8} \\
\quad \texttt{SACNet} & 70.06\,\tiny{$\pm$\,3.5} & 80.54\,\tiny{$\pm$\,2.5} & 91.93\,\tiny{$\pm$\,1.3} & 92.76\,\tiny{$\pm$\,1.3} & 97.00\,\tiny{$\pm$\,0.2} & 95.35\,\tiny{$\pm$\,0.4} & 95.10\,\tiny{$\pm$\,1.4} & 95.66\,\tiny{$\pm$\,0.3} \\
\quad \texttt{HybridSN} & 68.83\,\tiny{$\pm$\,2.0} & 80.85\,\tiny{$\pm$\,0.6} & 90.73\,\tiny{$\pm$\,4.2} & 92.16\,\tiny{$\pm$\,1.5} & 95.04\,\tiny{$\pm$\,0.9} & 93.00\,\tiny{$\pm$\,1.0} & 95.11\,\tiny{$\pm$\,1.1} & 95.43\,\tiny{$\pm$\,1.1} \\
\addlinespace[2pt]
\multicolumn{9}{@{}l}{\textit{Transformer}} \\[1pt]
\quad \texttt{MVAHN} & 82.71\,\tiny{$\pm$\,5.4} & 90.76\,\tiny{$\pm$\,2.5} & 97.57\,\tiny{$\pm$\,0.5} & 96.88\,\tiny{$\pm$\,0.9} & 98.17\,\tiny{$\pm$\,0.6} & 96.83\,\tiny{$\pm$\,1.2} & 97.48\,\tiny{$\pm$\,0.8} & 96.58\,\tiny{$\pm$\,0.7} \\
\quad \texttt{3DConvSST} & 81.79\,\tiny{$\pm$\,3.5} & 90.50\,\tiny{$\pm$\,1.5} & 96.05\,\tiny{$\pm$\,2.8} & 96.66\,\tiny{$\pm$\,1.7} & 97.45\,\tiny{$\pm$\,0.7} & 96.43\,\tiny{$\pm$\,0.1} & 96.19\,\tiny{$\pm$\,0.8} & 95.83\,\tiny{$\pm$\,0.9} \\
\quad \texttt{CTMixer} & \textbf{84.91\,\tiny{$\pm$\,1.9}} & 91.74\,\tiny{$\pm$\,1.3} & 97.07\,\tiny{$\pm$\,0.3} & 97.07\,\tiny{$\pm$\,0.5} & 96.63\,\tiny{$\pm$\,0.4} & 96.91\,\tiny{$\pm$\,0.2} & 92.58\,\tiny{$\pm$\,0.6} & 93.40\,\tiny{$\pm$\,0.4} \\
\quad \texttt{MMFormer} & 80.15\,\tiny{$\pm$\,4.5} & 88.25\,\tiny{$\pm$\,2.9} & 94.44\,\tiny{$\pm$\,3.2} & 95.81\,\tiny{$\pm$\,1.6} & \textbf{98.84\,\tiny{$\pm$\,0.2}} & \textbf{98.41\,\tiny{$\pm$\,0.3}} & 97.36\,\tiny{$\pm$\,0.3} & 96.93\,\tiny{$\pm$\,0.5} \\
\quad \texttt{DSFormer} & 80.50\,\tiny{$\pm$\,2.8} & 88.82\,\tiny{$\pm$\,0.8} & 96.50\,\tiny{$\pm$\,2.1} & 97.53\,\tiny{$\pm$\,0.6} & 98.20\,\tiny{$\pm$\,0.1} & 97.47\,\tiny{$\pm$\,0.9} & 95.49\,\tiny{$\pm$\,2.3} & 95.36\,\tiny{$\pm$\,1.4} \\
\quad \texttt{SSFTTNet} & 81.30\,\tiny{$\pm$\,3.1} & 89.47\,\tiny{$\pm$\,2.3} & 95.23\,\tiny{$\pm$\,1.9} & 96.03\,\tiny{$\pm$\,1.2} & 97.46\,\tiny{$\pm$\,1.0} & 97.00\,\tiny{$\pm$\,1.3} & 96.68\,\tiny{$\pm$\,0.8} & 96.62\,\tiny{$\pm$\,1.3} \\
\quad \texttt{MorphFormer} & 79.46\,\tiny{$\pm$\,3.6} & 88.77\,\tiny{$\pm$\,2.1} & 96.82\,\tiny{$\pm$\,0.4} & 96.63\,\tiny{$\pm$\,0.4} & 97.27\,\tiny{$\pm$\,0.1} & 96.50\,\tiny{$\pm$\,0.3} & 95.74\,\tiny{$\pm$\,1.0} & 95.56\,\tiny{$\pm$\,0.7} \\
\quad \texttt{MASSFormer} & 80.27\,\tiny{$\pm$\,2.7} & 89.37\,\tiny{$\pm$\,1.8} & 92.53\,\tiny{$\pm$\,3.8} & 94.29\,\tiny{$\pm$\,2.5} & 97.87\,\tiny{$\pm$\,0.4} & 97.46\,\tiny{$\pm$\,0.6} & 97.31\,\tiny{$\pm$\,0.5} & 96.64\,\tiny{$\pm$\,0.9} \\
\quad \texttt{FAHM} & 76.08\,\tiny{$\pm$\,6.0} & 86.52\,\tiny{$\pm$\,3.6} & 97.15\,\tiny{$\pm$\,0.5} & 96.21\,\tiny{$\pm$\,1.0} & 98.29\,\tiny{$\pm$\,0.5} & 97.67\,\tiny{$\pm$\,0.9} & 96.42\,\tiny{$\pm$\,2.0} & 96.33\,\tiny{$\pm$\,1.4} \\
\quad \texttt{GSCViT} & 77.33\,\tiny{$\pm$\,1.9} & 88.34\,\tiny{$\pm$\,1.3} & 95.55\,\tiny{$\pm$\,2.3} & 96.94\,\tiny{$\pm$\,0.6} & 98.16\,\tiny{$\pm$\,0.2} & 97.52\,\tiny{$\pm$\,0.7} & 96.59\,\tiny{$\pm$\,0.6} & 96.41\,\tiny{$\pm$\,1.1} \\
\quad \texttt{MFT} & 72.08\,\tiny{$\pm$\,3.9} & 83.65\,\tiny{$\pm$\,3.5} & 95.59\,\tiny{$\pm$\,1.2} & 95.52\,\tiny{$\pm$\,1.1} & 96.33\,\tiny{$\pm$\,1.8} & 94.74\,\tiny{$\pm$\,2.0} & 95.40\,\tiny{$\pm$\,1.6} & 94.93\,\tiny{$\pm$\,1.7} \\
\quad \texttt{SpectralFormer} & 74.18\,\tiny{$\pm$\,2.8} & 85.05\,\tiny{$\pm$\,2.5} & 91.69\,\tiny{$\pm$\,0.8} & 90.71\,\tiny{$\pm$\,0.7} & 97.16\,\tiny{$\pm$\,0.8} & 95.01\,\tiny{$\pm$\,2.1} & 96.24\,\tiny{$\pm$\,1.2} & 95.50\,\tiny{$\pm$\,0.7} \\
\quad \texttt{GAHT} & 82.14\,\tiny{$\pm$\,1.6} & 90.30\,\tiny{$\pm$\,1.4} & 91.77\,\tiny{$\pm$\,2.8} & 94.34\,\tiny{$\pm$\,0.9} & 92.62\,\tiny{$\pm$\,6.2} & 93.89\,\tiny{$\pm$\,3.0} & 88.88\,\tiny{$\pm$\,0.9} & 90.01\,\tiny{$\pm$\,1.4} \\
\quad \texttt{HSIC\_SClusterFormer} & 70.28\,\tiny{$\pm$\,3.3} & 81.58\,\tiny{$\pm$\,3.3} & 90.36\,\tiny{$\pm$\,1.4} & 92.74\,\tiny{$\pm$\,3.5} & 94.11\,\tiny{$\pm$\,1.7} & 93.27\,\tiny{$\pm$\,1.2} & 93.91\,\tiny{$\pm$\,3.9} & 92.62\,\tiny{$\pm$\,5.0} \\
\quad \texttt{S2Gformer} & 61.71\,\tiny{$\pm$\,1.9} & 77.32\,\tiny{$\pm$\,1.2} & 74.53\,\tiny{$\pm$\,7.4} & 82.39\,\tiny{$\pm$\,2.9} & 95.86\,\tiny{$\pm$\,0.7} & 93.66\,\tiny{$\pm$\,2.4} & 96.17\,\tiny{$\pm$\,1.2} & 94.70\,\tiny{$\pm$\,1.2} \\
\quad \texttt{DBCTNet} & 32.82\,\tiny{$\pm$\,6.8} & 33.87\,\tiny{$\pm$\,5.1} & 42.29\,\tiny{$\pm$\,20.7} & 44.97\,\tiny{$\pm$\,18.6} & 51.67\,\tiny{$\pm$\,6.9} & 44.71\,\tiny{$\pm$\,20.0} & 55.87\,\tiny{$\pm$\,8.2} & 51.87\,\tiny{$\pm$\,7.3} \\
\addlinespace[2pt]
\multicolumn{9}{@{}l}{\textit{Mamba / SSM}} \\[1pt]
\quad \texttt{R2Mamba} & 83.74\,\tiny{$\pm$\,2.3} & 91.28\,\tiny{$\pm$\,1.2} & 96.57\,\tiny{$\pm$\,1.7} & 96.89\,\tiny{$\pm$\,1.5} & 97.90\,\tiny{$\pm$\,0.6} & 98.04\,\tiny{$\pm$\,0.5} & \textbf{98.07\,\tiny{$\pm$\,0.4}} & \textbf{97.60\,\tiny{$\pm$\,0.1}} \\
\quad \texttt{MambaLG} & 82.47\,\tiny{$\pm$\,4.0} & 90.24\,\tiny{$\pm$\,1.1} & 97.21\,\tiny{$\pm$\,1.0} & 96.79\,\tiny{$\pm$\,0.6} & 97.60\,\tiny{$\pm$\,0.6} & 97.24\,\tiny{$\pm$\,1.1} & 97.53\,\tiny{$\pm$\,0.3} & 96.64\,\tiny{$\pm$\,0.6} \\
\quad \texttt{IGroupSS-Mamba} & 81.08\,\tiny{$\pm$\,0.9} & 90.35\,\tiny{$\pm$\,0.9} & 97.12\,\tiny{$\pm$\,0.7} & 97.65\,\tiny{$\pm$\,0.3} & 98.27\,\tiny{$\pm$\,0.0} & 98.26\,\tiny{$\pm$\,0.1} & 96.92\,\tiny{$\pm$\,0.5} & 96.72\,\tiny{$\pm$\,0.4} \\
\quad \texttt{S2Mamba} & 78.12\,\tiny{$\pm$\,3.5} & 88.10\,\tiny{$\pm$\,0.9} & 97.04\,\tiny{$\pm$\,0.7} & 97.63\,\tiny{$\pm$\,0.2} & 98.25\,\tiny{$\pm$\,0.1} & 97.94\,\tiny{$\pm$\,0.9} & 97.71\,\tiny{$\pm$\,0.7} & 97.09\,\tiny{$\pm$\,0.8} \\
\quad \texttt{HyperMamba} & 77.69\,\tiny{$\pm$\,1.6} & 87.59\,\tiny{$\pm$\,2.1} & 94.96\,\tiny{$\pm$\,1.0} & 95.84\,\tiny{$\pm$\,1.3} & 97.25\,\tiny{$\pm$\,0.4} & 97.60\,\tiny{$\pm$\,0.4} & 97.83\,\tiny{$\pm$\,0.3} & 97.58\,\tiny{$\pm$\,0.5} \\
\quad \texttt{PHDMamba} & 76.86\,\tiny{$\pm$\,3.2} & 87.38\,\tiny{$\pm$\,1.9} & 95.33\,\tiny{$\pm$\,0.8} & 96.42\,\tiny{$\pm$\,0.6} & 97.93\,\tiny{$\pm$\,0.6} & 97.45\,\tiny{$\pm$\,1.1} & 96.52\,\tiny{$\pm$\,1.1} & 96.40\,\tiny{$\pm$\,0.7} \\
\quad \texttt{MambaHSI\_Plus} & 77.18\,\tiny{$\pm$\,4.2} & 87.55\,\tiny{$\pm$\,1.8} & 94.25\,\tiny{$\pm$\,0.9} & 94.67\,\tiny{$\pm$\,1.0} & 98.13\,\tiny{$\pm$\,0.5} & 97.38\,\tiny{$\pm$\,0.6} & 96.40\,\tiny{$\pm$\,2.3} & 96.18\,\tiny{$\pm$\,1.6} \\
\quad \texttt{MambaMoE} & 75.89\,\tiny{$\pm$\,1.4} & 86.36\,\tiny{$\pm$\,1.6} & 94.43\,\tiny{$\pm$\,0.5} & 95.36\,\tiny{$\pm$\,0.9} & 97.70\,\tiny{$\pm$\,0.2} & 97.25\,\tiny{$\pm$\,0.9} & 95.98\,\tiny{$\pm$\,1.6} & 96.09\,\tiny{$\pm$\,1.5} \\
\quad \texttt{MiM} & 76.83\,\tiny{$\pm$\,9.3} & 87.56\,\tiny{$\pm$\,4.8} & 96.57\,\tiny{$\pm$\,1.0} & 97.21\,\tiny{$\pm$\,0.3} & 97.07\,\tiny{$\pm$\,1.0} & 95.98\,\tiny{$\pm$\,1.6} & 93.49\,\tiny{$\pm$\,1.4} & 92.54\,\tiny{$\pm$\,1.9} \\
\quad \texttt{MambaHSI} & 76.18\,\tiny{$\pm$\,2.8} & 86.22\,\tiny{$\pm$\,1.1} & 92.62\,\tiny{$\pm$\,2.0} & 93.77\,\tiny{$\pm$\,1.0} & 97.87\,\tiny{$\pm$\,0.2} & 96.94\,\tiny{$\pm$\,0.7} & 95.23\,\tiny{$\pm$\,1.2} & 94.47\,\tiny{$\pm$\,1.9} \\
\quad \texttt{EMamba} & 73.54\,\tiny{$\pm$\,2.0} & 86.28\,\tiny{$\pm$\,1.6} & 95.78\,\tiny{$\pm$\,0.9} & 96.56\,\tiny{$\pm$\,0.3} & 97.01\,\tiny{$\pm$\,0.5} & 96.72\,\tiny{$\pm$\,0.2} & 95.33\,\tiny{$\pm$\,0.3} & 94.35\,\tiny{$\pm$\,0.5} \\
\quad \texttt{HyPyraMamba} & 73.72\,\tiny{$\pm$\,2.8} & 84.88\,\tiny{$\pm$\,1.9} & 93.11\,\tiny{$\pm$\,1.4} & 93.89\,\tiny{$\pm$\,0.7} & 96.84\,\tiny{$\pm$\,1.3} & 96.53\,\tiny{$\pm$\,1.6} & 97.04\,\tiny{$\pm$\,0.7} & 95.96\,\tiny{$\pm$\,0.6} \\
\quad \texttt{FuzzySpectralMamba} & 73.23\,\tiny{$\pm$\,3.6} & 85.20\,\tiny{$\pm$\,2.2} & 95.74\,\tiny{$\pm$\,1.6} & 96.63\,\tiny{$\pm$\,0.4} & 96.72\,\tiny{$\pm$\,1.1} & 95.71\,\tiny{$\pm$\,1.1} & 94.85\,\tiny{$\pm$\,0.5} & 93.43\,\tiny{$\pm$\,0.2} \\
\quad \texttt{MLFMamba} & 76.62\,\tiny{$\pm$\,5.2} & 86.91\,\tiny{$\pm$\,2.1} & 86.16\,\tiny{$\pm$\,4.7} & 90.34\,\tiny{$\pm$\,0.9} & 97.86\,\tiny{$\pm$\,0.2} & 96.94\,\tiny{$\pm$\,0.9} & 96.68\,\tiny{$\pm$\,0.9} & 96.24\,\tiny{$\pm$\,0.7} \\
\quad \texttt{MorpMamba} & 64.30\,\tiny{$\pm$\,1.5} & 78.07\,\tiny{$\pm$\,0.9} & 88.97\,\tiny{$\pm$\,4.0} & 90.11\,\tiny{$\pm$\,2.4} & 95.51\,\tiny{$\pm$\,0.4} & 94.68\,\tiny{$\pm$\,0.7} & 93.06\,\tiny{$\pm$\,0.7} & 93.87\,\tiny{$\pm$\,0.7} \\
\quad \texttt{SSMamba} & 63.12\,\tiny{$\pm$\,8.3} & 77.15\,\tiny{$\pm$\,5.8} & 85.58\,\tiny{$\pm$\,8.1} & 90.04\,\tiny{$\pm$\,3.5} & 97.07\,\tiny{$\pm$\,1.0} & 97.57\,\tiny{$\pm$\,0.5} & 92.74\,\tiny{$\pm$\,4.1} & 93.21\,\tiny{$\pm$\,2.3} \\
\quad \texttt{MHSSMamba} & 59.66\,\tiny{$\pm$\,7.5} & 72.77\,\tiny{$\pm$\,7.1} & 76.71\,\tiny{$\pm$\,2.2} & 78.58\,\tiny{$\pm$\,0.7} & 91.39\,\tiny{$\pm$\,4.3} & 84.28\,\tiny{$\pm$\,7.2} & 94.49\,\tiny{$\pm$\,1.7} & 93.05\,\tiny{$\pm$\,1.7} \\
\quad \texttt{GraphMamba} & 26.60\,\tiny{$\pm$\,40.1} & 31.53\,\tiny{$\pm$\,43.8} & 94.87\,\tiny{$\pm$\,4.1} & 96.11\,\tiny{$\pm$\,1.6} & 96.88\,\tiny{$\pm$\,1.5} & 97.14\,\tiny{$\pm$\,0.6} & 96.32\,\tiny{$\pm$\,0.9} & 93.95\,\tiny{$\pm$\,4.2} \\
\quad \texttt{ConvVitMamba} & 52.39\,\tiny{$\pm$\,4.0} & 67.47\,\tiny{$\pm$\,4.3} & 77.59\,\tiny{$\pm$\,22.5} & 83.49\,\tiny{$\pm$\,9.5} & 86.75\,\tiny{$\pm$\,12.1} & 84.89\,\tiny{$\pm$\,10.4} & 95.41\,\tiny{$\pm$\,1.1} & 93.78\,\tiny{$\pm$\,1.9} \\
\quad \texttt{WaveMamba} & 31.98\,\tiny{$\pm$\,4.1} & 27.02\,\tiny{$\pm$\,4.0} & 48.05\,\tiny{$\pm$\,10.7} & 39.64\,\tiny{$\pm$\,10.6} & 74.53\,\tiny{$\pm$\,10.2} & 67.45\,\tiny{$\pm$\,3.3} & 80.02\,\tiny{$\pm$\,2.1} & 73.89\,\tiny{$\pm$\,3.6} \\
\addlinespace[2pt]
\multicolumn{9}{@{}l}{\textit{Graph / GCN}} \\[1pt]
\quad \texttt{MS2GCAN} & 84.15\,\tiny{$\pm$\,7.5} & 91.08\,\tiny{$\pm$\,4.3} & 94.23\,\tiny{$\pm$\,3.3} & 93.09\,\tiny{$\pm$\,4.9} & 98.33\,\tiny{$\pm$\,1.1} & 98.36\,\tiny{$\pm$\,0.4} & 96.88\,\tiny{$\pm$\,0.2} & 95.50\,\tiny{$\pm$\,1.6} \\
\quad \texttt{GTCFN} & 77.68\,\tiny{$\pm$\,2.8} & 88.28\,\tiny{$\pm$\,1.6} & 96.74\,\tiny{$\pm$\,1.2} & 96.61\,\tiny{$\pm$\,1.0} & 97.35\,\tiny{$\pm$\,0.1} & 95.99\,\tiny{$\pm$\,1.0} & 93.57\,\tiny{$\pm$\,1.2} & 94.17\,\tiny{$\pm$\,1.3} \\
\quad \texttt{MCTGCL} & 72.32\,\tiny{$\pm$\,6.1} & 85.63\,\tiny{$\pm$\,2.9} & 87.27\,\tiny{$\pm$\,3.4} & 91.47\,\tiny{$\pm$\,3.6} & 97.39\,\tiny{$\pm$\,0.5} & 96.52\,\tiny{$\pm$\,0.9} & 95.46\,\tiny{$\pm$\,2.0} & 94.90\,\tiny{$\pm$\,1.8} \\
\quad \texttt{GraphGST} & 71.48\,\tiny{$\pm$\,4.2} & 85.26\,\tiny{$\pm$\,1.4} & 91.70\,\tiny{$\pm$\,2.2} & 91.60\,\tiny{$\pm$\,2.9} & 92.63\,\tiny{$\pm$\,4.5} & 91.50\,\tiny{$\pm$\,3.4} & 89.33\,\tiny{$\pm$\,4.0} & 78.23\,\tiny{$\pm$\,5.2} \\
\addlinespace[2pt]
\multicolumn{9}{@{}l}{\textit{KAN}} \\[1pt]
\quad \texttt{HyperKAN} & 79.00\,\tiny{$\pm$\,3.0} & 88.76\,\tiny{$\pm$\,1.5} & 96.15\,\tiny{$\pm$\,1.1} & 96.37\,\tiny{$\pm$\,1.0} & 97.45\,\tiny{$\pm$\,0.3} & 97.72\,\tiny{$\pm$\,0.0} & 94.38\,\tiny{$\pm$\,0.8} & 94.97\,\tiny{$\pm$\,0.9} \\
\quad \texttt{HSIConvKAN} & 59.65\,\tiny{$\pm$\,5.5} & 74.89\,\tiny{$\pm$\,2.6} & 86.48\,\tiny{$\pm$\,2.4} & 87.33\,\tiny{$\pm$\,1.1} & 92.47\,\tiny{$\pm$\,1.2} & 92.12\,\tiny{$\pm$\,1.0} & 86.31\,\tiny{$\pm$\,2.2} & 87.46\,\tiny{$\pm$\,1.7} \\
\addlinespace[2pt]
\multicolumn{9}{@{}l}{\textit{Self-supervised}} \\[1pt]
\quad \texttt{HSIMAE} & 65.81\,\tiny{$\pm$\,2.4} & 79.96\,\tiny{$\pm$\,1.6} & 84.04\,\tiny{$\pm$\,3.6} & 86.67\,\tiny{$\pm$\,2.8} & 95.84\,\tiny{$\pm$\,1.2} & 92.03\,\tiny{$\pm$\,4.1} & 95.94\,\tiny{$\pm$\,0.6} & 95.50\,\tiny{$\pm$\,0.3} \\
\quad \texttt{LFSMIM} & 60.20\,\tiny{$\pm$\,3.2} & 74.54\,\tiny{$\pm$\,1.3} & 87.86\,\tiny{$\pm$\,4.1} & 88.71\,\tiny{$\pm$\,3.0} & 95.02\,\tiny{$\pm$\,1.5} & 95.85\,\tiny{$\pm$\,0.6} & 92.52\,\tiny{$\pm$\,2.0} & 92.69\,\tiny{$\pm$\,1.3} \\
\quad \texttt{HSIC\_FM} & 31.93\,\tiny{$\pm$\,25.3} & 47.35\,\tiny{$\pm$\,27.2} & 70.28\,\tiny{$\pm$\,7.6} & 72.50\,\tiny{$\pm$\,4.5} & 78.98\,\tiny{$\pm$\,5.4} & 80.20\,\tiny{$\pm$\,6.5} & 34.43\,\tiny{$\pm$\,10.6} & 21.41\,\tiny{$\pm$\,4.6} \\
\addlinespace[2pt]
\bottomrule
\end{tabular}
\end{table}

Four things are visible in that table which the aggregates hide.

First, the ordering of architectures changes with the scene. SSRN leads Pavia University
at 97.81\% overall accuracy and leads none of the other three; the highest values on
Indian Pines, WHU Hi LongKou and Nili Fossae belong to CTMixer, MMFormer and R2Mamba
respectively, drawn from two different families. Rank correlation across scenes is
positive but far from unity.

Second, the gap between overall and average accuracy is a scene property rather than a
model property. On Indian Pines, whose imbalance ratio is $123\times$, average accuracy
exceeds overall accuracy for 54 of the 55 models, because drawing a fixed 10 training
pixels from every class, the reduced budget this scene requires (\Cref{sec:classstats}),
gives a minority class a training share far above its share of the test set. On
WHU Hi LongKou, at $22\times$, the inequality holds for only 10 of the 55. This is the
behaviour \Cref{eq:metrics} predicts from the imbalance ratios in \Cref{tab:datasets},
and it is why the framework reports both metrics rather than either alone.

Third, the seeded standard deviations are informative in themselves. On Pavia University
the strongest models vary by a few tenths of a point across seeds, while on Indian Pines
the same models vary by three to five points. A single-run comparison on Indian Pines can
therefore reverse an ordering purely by seed, which is a concrete argument for repeated
runs rather than a methodological preference.

Fourth, Nili Fossae shows that terrestrial architectures transfer to planetary data
without modification. Mean accuracies there are comparable to the easier terrestrial
scenes despite 425 spectral bands, only nine classes and mineralogical rather than
land-cover semantics, and the paradigm ordering is broadly preserved. That comparison was
not previously available without reimplementing every model against a CRISM loader.

\subsection{Accuracy against computational cost}\label{sec:complexity}

\begin{figure}[t]
\centering
\includegraphics[width=\linewidth]{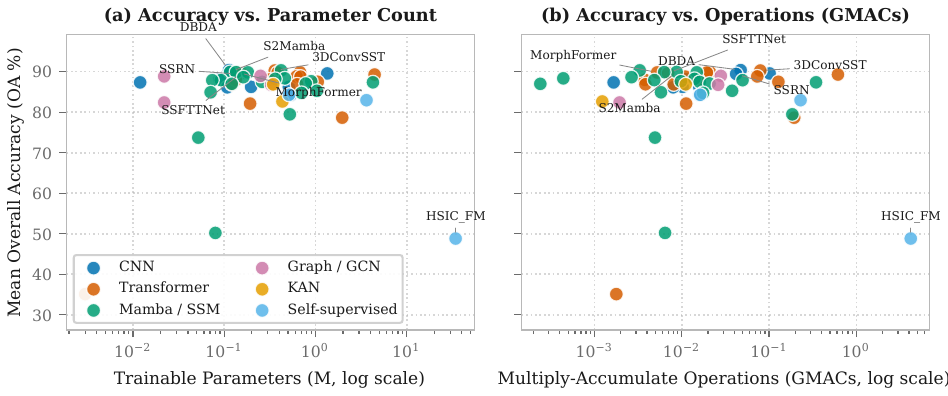}
\caption{Mean overall accuracy against (a) trainable parameters and (b) multiply-accumulate
operations (GMACs), both on logarithmic axes and both measured under one fixed probe
input. Annotations mark the architectures on the accuracy and cost frontier.}
\label{fig:complexity}
\end{figure}

Under this protocol, scale and accuracy are weakly negatively associated:
$r = \ParamOACorr$ against parameter count and $r = \FlopOACorr$ against operation count.
The frontier is populated by compact models. DBDA reaches 90.37\% with 0.113\,M
parameters and 0.047\,GMACs; S2Mamba reaches 89.91\% with 0.116\,M parameters and under
0.01\,GMACs; MorphFormer and SSFTTNet reach roughly 89.7\% with similar footprints. At the
other end, the foundation-scale HSIC\_FM uses 34.2\,M parameters and 4.20\,GMACs to reach
48.82\%, and MMFormer, the largest transformer in the collection at 4.44\,M parameters,
attains 89.28\%, inside the range set by models thirty times smaller. For onboard or edge
deployment this is the practical result to carry forward: compact architectures deliver
essentially the accuracy of the largest models in this benchmark at two orders of
magnitude less compute.

\subsection{Visual analysis}\label{sec:qualitative}

\begin{figure}[p]
\centering
\includegraphics[width=0.96\linewidth]{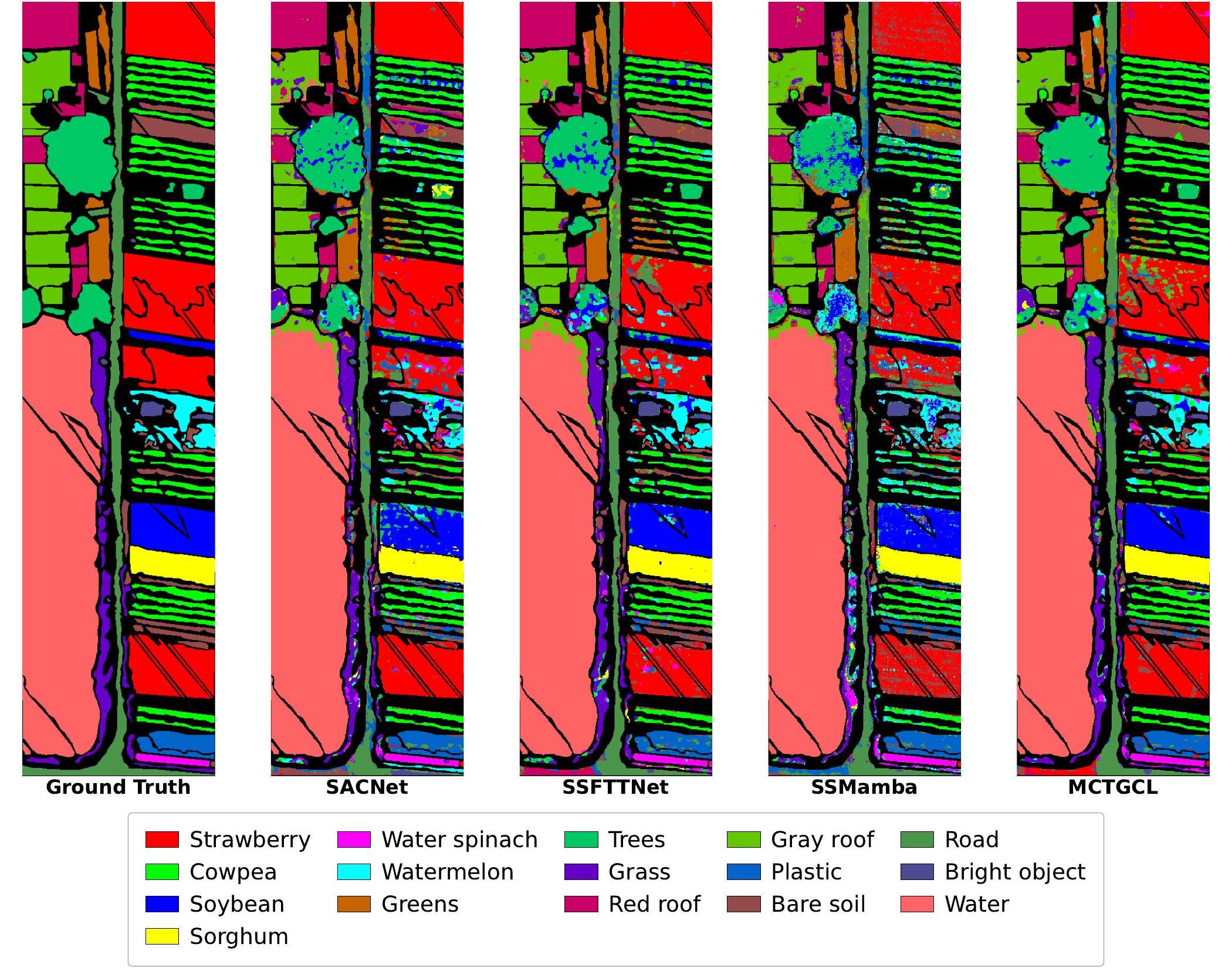}\\[3pt]
{\small\textbf{(a)} WHU Hi HanChuan (UAV agricultural scene, 16 crop classes)}\\[12pt]
\includegraphics[width=0.96\linewidth]{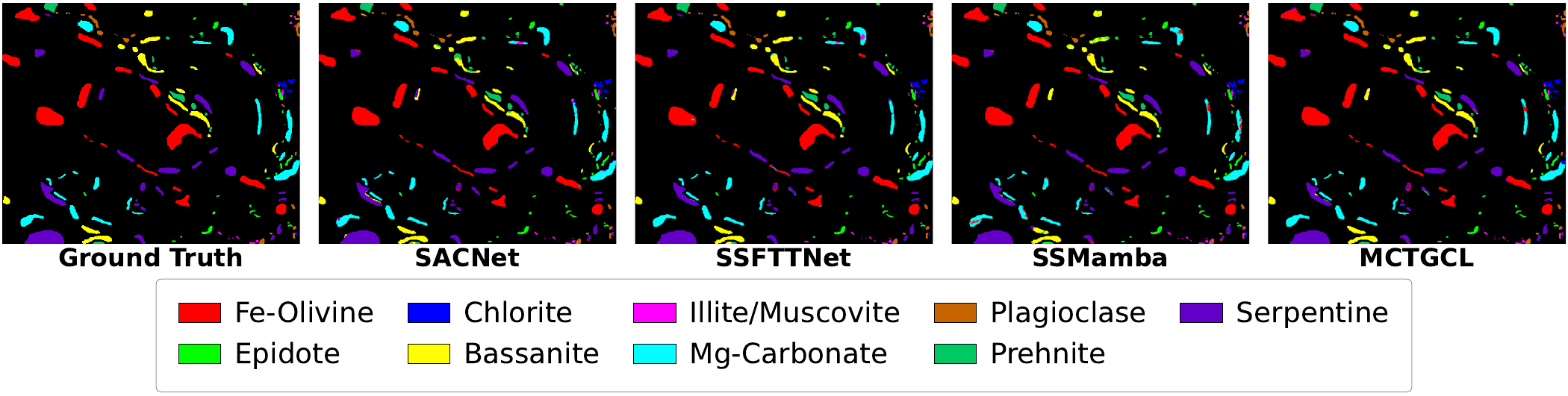}\\[3pt]
{\small\textbf{(b)} Nili Fossae (Mars CRISM planetary orbital scene, 9 mineralogical classes)}\\[6pt]
\caption{Arranged classification maps generated directly by the framework. (a) WHU Hi HanChuan, a UAV agricultural strip with narrow parcel boundaries. (b) Nili Fossae, a Mars CRISM scene with diffuse mineralogical transitions. Each row displays the ground-truth map alongside representative convolutional (\texttt{SACNet}), transformer (\texttt{SSFTTNet}), state-space (\texttt{SSMamba}), and graph-based (\texttt{MCTGCL}) model predictions with comprehensive class palettes and legends.}
\label{fig:maps}
\end{figure}

Aggregate metrics hide spatial failure modes, which is why the framework assembles
multi-model map figures directly from completed runs, with the ground truth, a shared
colour map and a class legend read from the dataset registry.
\Cref{fig:maps} shows two scenes chosen for their contrast: WHU Hi HanChuan, whose
classes are narrow elongated crop strips with sharp boundaries, and Nili Fossae, whose
mineralogical boundaries are diffuse gradients.

Three characteristics are visible. Convolutional and hybrid transformer models
(\texttt{SACNet}, \texttt{SSFTTNet}) hold crop-row boundaries crisply, with little spill-over
across parcel edges. State-space models (\texttt{SSMamba}, \texttt{MambaHSI\_Plus}) smooth
interiors strongly and remove isolated misclassified pixels, but produce occasional
directional streaking near transitions, aligned with the scan trajectory. Graph models
(\texttt{MCTGCL}) yield piecewise-constant regions that are clean in open parcels and that
merge small features where superpixel boundaries do not follow class boundaries. On Nili
Fossae the ordering shifts: models that impose strong regional constancy over-segment
continuous mineralogical gradients, while \texttt{GAHT} and \texttt{FETNet} track them
more faithfully.

\subsection{Discussion}\label{sec:discussion}

Read together, the quantitative and visual evidence supports four statements about the
current state of hyperspectral classification, each of which is a claim about behaviour
under a shared protocol rather than about any individual publication.

\paragraph{The scene matters more than the architecture.}
The \SceneSpread{}-point spread across scenes against a \FamilySpread{}-point spread
across family means is the central number in this benchmark. A claimed improvement on Indian Pines and Pavia
University is weak evidence about behaviour on UAV, urban or planetary data, so the
marginal value of a new architecture measured on the classical scenes alone is small
against what a broader collection reveals. That is an argument for breadth of evaluation,
and it is the argument the framework is built to make cheap to act on.

\paragraph{Established inductive biases remain highly competitive under the reference small-sample protocol.}
Under 30 labelled samples per class a scene supplies only a few hundred training patches,
and a large, loosely constrained model has little to fit them with beyond its priors. That is why
local weight sharing, structured spectral grouping and bounded attention behave as
regularisers here, why the fitted year effect is small, and why the complexity analysis
(\Cref{sec:complexity}) points the same way: the top of the accuracy distribution is occupied by
sub-1\,M-parameter models rather than the largest architectures in the catalogue. None of it shows
that capacity is counterproductive in general; it shows that this regime, which is the one most
hyperspectral annotation budgets impose, rewards architectural economy. Whether the ordering
survives at 200 samples per class is one configuration key away and is the natural next experiment.

\paragraph{Serialisation is the open question for state-space models.}
The state-space family reaches the accuracy of the best transformers at its top end while
being the most dispersed family overall, and the streaking in \Cref{fig:maps} points at
why. A selective scan needs a one-dimensional ordering of a two-dimensional
neighbourhood, and at an $11\times11$ window the sequence is only 121 tokens long, so the
asymptotic advantage of linear-time scanning has little room to express itself while the
cost of choosing an ordering remains. Variation within the family tracks how each design
handles that choice more closely than it tracks publication date, which suggests scan
design rather than scan efficiency is where the remaining gains are.

\paragraph{Aggregates conceal failure modes that per-scene reporting exposes.}
Aggregate family means can be skewed by a single weak member (DBCTNet among the
transformers, HSIC\_FM among the self-supervised models), even when the same family
contains some of the most consistent architectures in the collection. Neither figure
should be read as a verdict on the architecture. \texttt{DBCTNet} instantiates to
0.003\,M parameters under the shared constructor contract, two orders of magnitude below
its published configuration, which points at how its width is derived from the input
rather than at the design itself; \texttt{HSIC\_FM} is a foundation-scale model run here
without the pretraining stage that gives it its representations. Both are reported
because the framework reports what it measured, and both are cases where the per-model
record in \Cref{app:models}, together with the deviation note carried in each model file,
is what a reader needs rather than the family mean.
A benchmark that reported only aggregate metrics would licence misleading conclusions
about architectural paradigms. \Cref{tab:perscene}, the dispersion measures of \Cref{sec:families} and the
maps of \Cref{fig:maps} each recover information the aggregate discards, which is why the
framework generates all of them from the same run records rather than reducing a sweep to
a single table.

% =====================================================================
%  sec/5b_leakage.tex
% =====================================================================
\subsection{Spatially disjoint partitioning and the guard band}\label{sec:disjoint}

The reference protocol evaluates models under the standard per-class split for
comparability with the bulk of published results. For evaluation under spatial shift, the
framework draws a spatially disjoint partition instead (\Cref{sec:splitting}): training,
validation and test patches occupy separate regions of the scene, and the guard band of
\cref{eq:guard} removes every validation or test window that would share a pixel with a
training window. This section reports how that partition behaves across the complete
\NumDatasets{}-scene collection.

\paragraph{The disjoint protocol on the full collection.}
\Cref{tab:disjointsplit} reports the disjoint split drawn at the default 50/30/20 targets
on all \NumDatasets{} scenes. Three observations follow from it. First, assigning patches by their
centre pixel alone does not separate them: between 6.7\% (Chikusei) and 81.6\% (Indian
Pines) of the validation and test patches share pixels with a training window, with a
median of 23.5\%. The overlap exceeds a quarter of the evaluation set on 11 scenes and half
of it on four (Augsburg, Loukia, MUUFL and Indian Pines), all scenes whose classes are
small, fragmented parcels, so that most of each region lies within a window's reach of a
region boundary. Without the guard band, a disjoint split of Indian Pines would be disjoint
in name only; with it, those patches are removed rather than evaluated. Second, the targets
are approximate: whole components are assigned, so the realised test share falls to 10.4\%
on Salinas and 11.1\% on Pavia Centre against a target of 20\%, and individual classes land
up to 22.8 percentage points from their targets (Salinas). Third, between 92.3\% (MUUFL) and
100\% (Chikusei) of the labelled pixels can centre a window; the remainder lie within five
pixels of the image border and are excluded under every protocol, including the showcase
benchmark, which is worth stating whenever results are compared against an implementation
that pads the scene.

% Transcribed from docs/DISJOINT_SPLIT_CLASSES.md (ratio mode, 50/30/20, P=11, stride 1, seed 0).
\begin{table}[t]
\centering
\caption{The disjoint protocol drawn on every scene at the default 50/30/20 targets ($P=11$, stride 1, seed 0), with the per-class breakdown in \texttt{docs/DISJOINT\_SPLIT\_CLASSES.md}. \emph{Kept} is the share of labelled pixels that can centre an $11\times11$ window; the rest lie within five pixels of the border and are never used, under any protocol. \emph{Split} gives the realised train/validation/test shares of the kept patches, and the next two columns how far the individual classes land from their targets, averaged over classes and at the worst class, in percentage points. \emph{Overlap} is the share of validation and test patches whose window shares at least one pixel with a training window when patches are assigned by their centre pixel alone; these are exactly the patches the guard band of \cref{eq:guard} removes.}
\label{tab:disjointsplit}
\footnotesize
\setlength{\tabcolsep}{4pt}
\begin{tabular}{@{}lrrrcrrr@{}}
\toprule
& & & & & \multicolumn{2}{c}{Off target (pp)} & \\
\cmidrule(lr){6-7}
Scene & Cls & Labelled px & Kept (\%) & Split (\%) & Mean & Worst & Overlap (\%) \\
\midrule
\addlinespace[2pt] \multicolumn{8}{@{}l}{\textit{Airborne} (14)} \\[1pt]
Augsburg & 7 & 78,294 & 94.4 & 49.5/31.9/18.6 & 2.6 & 10.8 & 58.9 \\
Berlin & 8 & 464,671 & 99.7 & 50.7/30.1/19.2 & 1.1 & 4.6 & 31.9 \\
Chikusei & 19 & 77,592 & 100.0 & 47.2/32.8/20.0 & 5.1 & 16.2 & 6.7 \\
Dioni & 12 & 20,024 & 98.9 & 49.8/29.6/20.6 & 1.8 & 6.8 & 27.6 \\
Houston 2013 & 15 & 15,029 & 99.7 & 50.2/29.8/20.0 & 0.7 & 3.0 & 41.7 \\
Houston 2018 & 20 & 2,018,910 & 99.1 & 49.9/30.1/20.0 & 2.4 & 14.6 & 18.6 \\
Indian Pines & 16 & 10,249 & 94.9 & 50.4/32.7/16.8 & 6.6 & 17.5 & 81.6 \\
KSC & 13 & 5,211 & 97.9 & 45.4/34.0/20.6 & 6.8 & 17.4 & 19.7 \\
Loukia & 14 & 13,503 & 96.2 & 48.8/30.6/20.7 & 3.1 & 12.5 & 66.5 \\
MUUFL & 11 & 53,687 & 92.3 & 50.6/30.0/19.4 & 4.1 & 15.6 & 80.3 \\
Pavia Centre & 9 & 148,152 & 98.3 & 54.1/34.8/11.1 & 4.2 & 19.4 & 25.2 \\
Pavia University & 9 & 42,776 & 94.5 & 42.2/38.0/19.8 & 7.2 & 15.2 & 34.9 \\
Salinas & 16 & 54,129 & 96.2 & 52.0/37.6/10.4 & 10.2 & 22.8 & 32.7 \\
Trento & 6 & 30,214 & 99.0 & 51.3/32.2/16.4 & 6.8 & 18.2 & 15.8 \\
\addlinespace[2pt] \multicolumn{8}{@{}l}{\textit{Spaceborne} (1)} \\[1pt]
Botswana & 14 & 3,248 & 99.7 & 47.2/31.6/21.2 & 4.9 & 11.3 & 24.8 \\
\addlinespace[2pt] \multicolumn{8}{@{}l}{\textit{Unmanned aerial vehicle} (6)} \\[1pt]
Pingan & 10 & 1,140,937 & 98.2 & 46.0/31.6/22.5 & 5.5 & 12.5 & 9.1 \\
Qingyun & 6 & 954,893 & 98.1 & 51.5/31.3/17.3 & 4.8 & 11.2 & 9.8 \\
Tangdaowan & 18 & 557,366 & 98.3 & 50.3/31.1/18.6 & 5.9 & 13.6 & 8.0 \\
WHU-Hi-HanChuan & 16 & 257,530 & 96.1 & 50.4/32.1/17.4 & 3.8 & 10.8 & 28.6 \\
WHU-Hi-HongHu & 22 & 386,693 & 97.9 & 49.6/31.4/19.0 & 5.6 & 17.2 & 14.7 \\
WHU-Hi-LongKou & 9 & 204,542 & 95.9 & 49.6/32.7/17.7 & 5.1 & 12.1 & 22.1 \\
\addlinespace[2pt] \multicolumn{8}{@{}l}{\textit{Planetary orbital} (3)} \\[1pt]
Holden & 6 & 20,090 & 95.8 & 47.9/31.3/20.9 & 3.1 & 6.7 & 12.7 \\
Nili Fossae & 9 & 26,710 & 95.2 & 49.4/30.6/20.0 & 4.2 & 12.3 & 18.2 \\
Utopia & 9 & 17,338 & 94.4 & 51.4/30.5/18.0 & 2.9 & 8.2 & 21.0 \\
\bottomrule
\end{tabular}
\end{table}

\subsection{Scope and Interpretation of the Showcase Benchmark}\label{sec:scope}

The benchmark presented in this section is intended as a representative showcase of the Hyperspectral-Image-Models framework rather than as a fixed definition of the experimental space supported by the library. The reported results use a controlled reference configuration of $11\times11$ spatial patches, 30 PCA components, and a nominal budget of 30 training and 10 validation samples per class, with the reduced 10/5 allocation used for Indian Pines. This configuration was selected to provide a common reference point across all \NumCatalogModels{} architectures and \NumDatasets{} scenes.

The framework itself is not restricted to this setting. Patch dimensions, spectral preprocessing, sampling budgets, partitioning strategies, optimization settings, and the number of repeated runs are exposed through the declarative configuration system. The library also supports spatially disjoint evaluation as an alternative to the standard class-balanced sampling used in the showcase benchmark. Multimodal architectures such as \texttt{EMamba} are evaluated here in a single-modality setting to enable direct cross-paradigm comparison. Complexity is characterized through structural graph profiling rather than hardware-dependent wall-clock latency. These design choices define the scope of the showcase rather than restricting the underlying framework, allowing researchers to construct task-specific experiments without modifying the core execution infrastructure.

Accordingly, the quantitative findings in this section should be interpreted as observations under the stated reference configuration rather than as universal rankings of architectural families. The primary value of the showcase is to demonstrate that heterogeneous published architectures can be executed, compared, and analysed within one reproducible environment, while the framework provides the configuration space needed to investigate other experimental regimes.

\paragraph{A protocol choice: overlapping windows under random sampling.}
The reference protocol draws training, validation and test pixels at random from the same
scene for comparability with the bulk of published results. With an $11\times11$ window, a
test patch whose centre lies within ten pixels of a training centre shares pixels with that
training patch, and within five pixels it contains that labelled training pixel itself
\citep{liang2017sampling}. Because of this window overlap, the absolute accuracies reported
here should be read as a common-protocol comparison between models rather than an estimate
of accuracy on spatially unseparated ground, and the gap to spatially separated evaluation
differs between architectures that use spatial context heavily and those that do not. The
spatially disjoint protocol with its guard band (\Cref{sec:splitting}) provides a
spatial-shift setting in which no evaluation window shares a pixel with a training window,
and \texttt{tools/disjoint\_audit.py} reports, from the ground truth alone, how large the
window overlap is under any partition before a model is trained (\Cref{sec:disjoint}).

\paragraph{A methodological limitation: scene-wide preprocessing fit.} As noted in \Cref{sec:datalayer}, min--max normalisation and PCA are fitted over the complete scene before the train/validation/test partition is drawn. Both transforms are label-agnostic, so no class information crosses the split; however, the fitted PCA basis and normalisation range are still computed using pixels that later fall inside the test partition, so the spectral coordinate system is not strictly independent of the test region. This is standard practice in the hyperspectral benchmarking literature we follow \citep{audebert2019deep} and does not affect the relative comparison between architectures reported here, since every model sees the identical transform. It is nonetheless a protocol choice a stricter benchmark could avoid by fitting normalisation and PCA on the training partition alone and applying the resulting transform to validation and test pixels. the configuration system in Hyperspectral-Image-Models supports implementing this train-only fitting mode without changing any model code; we flag it as a direction for a future revision of the benchmark rather than re-running the full \NumTotalRuns{}-run showcase under it here.

\FloatBarrier
% =====================================================================
%  sec/9_conclusion.tex
% =====================================================================
\section{Conclusion}

Hyperspectral-Image-Models was designed to make hyperspectral image classification rigorous, reproducible, and
comparable. It places \NumCatalogModels{} published architectures from diverse paradigms behind a
unified registry interface, executes them across \NumDatasets{} multiplatform scenes spanning airborne,
spaceborne, UAV, and planetary orbital acquisitions, and preserves full experimental provenance alongside
every emitted result. The showcase benchmark presented in this work, comprising \NumTotalEvaluations{}
model--scene evaluations over \NumTotalRuns{} seeded training runs, serves as an extensive demonstration of the
library's end-to-end capabilities and establishes a comprehensive, reproducible reference benchmark that researchers
can directly build upon without redundant reimplementation.

The empirical findings of this showcase benchmark provide key insights into current classification dynamics.
Scene difficulty accounts for far more variance than architectural choice; no single paradigm dominates
universally, with top performance distributed according to scene geometry rather than publication recency;
the historical accuracy gain over the past decade is modest relative to within-year dispersion; and in the
small-sample regime, several compact architectures rival the accuracy of substantially larger models,
including models more than two orders of magnitude larger in parameter count. We emphasize that while
these findings are measured under the showcase reference protocol ($11\times11$ spatial patches, 30 PCA
components, and a reference budget of 30 samples per class, reduced to 10 on Indian Pines), the
Hyperspectral-Image-Models framework natively supports arbitrary spatial patch dimensions, diverse spectral preprocessing
modes, continuous ratio splits, and spatially disjoint partitioning with a guard band that removes
evaluation windows overlapping any training window, together with an audit that measures this overlap
from the ground truth before any model is trained. Researchers can re-examine
and extend these frontiers across varying patch sizes, split ratios, and spatial isolation strategies
through simple configuration modifications. The framework and associated code are released under the
Apache~2.0 licence. The dataset collection is distributed with attribution, while individual source
datasets remain subject to their original licences and terms. We invite the community to accelerate
fair, transparent, and reproducible discovery in hyperspectral remote sensing.

\FloatBarrier

% =====================================================================
%  Back matter
% =====================================================================
\section*{Data and Code Availability}
The framework and associated code are openly available at
\href{https://github.com/Tanishq251/Hyperspectral-Image-Models}{\textcolor{repoblue}{\url{https://github.com/Tanishq251/Hyperspectral-Image-Models}}} under the Apache~2.0 licence, and the benchmark tables and comparison maps reported here are generated directly by the framework from those same completed runs.
The \NumDatasets{}-scene standardised dataset collection is curated with attribution at
\href{https://huggingface.co/datasets/Tanishq165/HSI_Datasets}{\textcolor{repoblue}{\url{https://huggingface.co/datasets/Tanishq165/HSI_Datasets}}}, while individual
source datasets remain subject to their original licences and terms. All run configurations,
generated tables and figure scripts are archived with the manuscript source.

\section*{Acknowledgements}
Every architecture in the framework is an adaptation or reimplementation of published
work, and architectural credit belongs to the original authors cited in
\Cref{tab:families} and \Cref{tab:models}. We thank the providers of the
\NumDatasets{} public hyperspectral scenes, including NASA JPL and the AVIRIS team,
NASA MRO CRISM, IEEE GRSS, Wuhan University, the University of Pavia, DLR and the
HyMap team, Ocean University of China, the University of Southern Mississippi, and the
HyRANK and Chikusei consortia.

{\footnotesize
\setlength{\bibsep}{0.2pt plus 0.1ex}
\begin{multicols}{2}
\bibliographystyle{unsrtnat}
\bibliography{ref}
\end{multicols}
}

\clearpage
\appendix
\section{The full model catalogue}\label{app:models}

\Cref{tab:models} lists every architecture the framework implements, grouped by
family and ordered by publication year, with the venue and the digital object
identifier of the original paper. The entries are transcribed from
\texttt{MODEL\_CATALOG} in \texttt{models/registry.py}, which is the same source
the repository's own documentation and generated tables read from, so the paper and
the code cannot disagree about what is implemented. The official reference
implementation of every model is recorded in \texttt{MODEL\_CATALOG} alongside the
identifier cited here.

One convention in that table is worth restating: a citation is taken from each
entry's \texttt{paper\_title} and identifier rather than from its
\texttt{full\_name}, which is a descriptive label maintained for display and is not
always the published title. The table records what the framework implements; it
reports no measurement of any kind, and none should be read into an architecture's
presence in it.

\begin{table}[p]
\centering
\caption{The complete model catalogue, grouped by family and ordered by publication year, transcribed from \texttt{MODEL\_CATALOG} in \texttt{models/registry.py} and \texttt{docs/MODELS.md}. Parameters and multiply-accumulate operations are measured by \texttt{model\_info.py} under one fixed probe input of shape $(1,1,30,P,P)$ at 16 classes, where $P$ is the patch column; they are structural properties, not performance measurements. Architectural credit for every entry belongs to its original authors.}
\label{tab:models}
\scriptsize
\setlength{\tabcolsep}{4pt}
\begin{tabular}{@{}llllrrl@{}}
\toprule
Model & Yr & Venue & Par. (M) & MACs (M) & $P$ & Paper \\
\midrule
\addlinespace[2pt] \multicolumn{7}{@{}l}{\textit{CNN} (10 implemented)} \\[1pt]
\quad \texttt{SSRN} & 2017 & TGRS & 0.136 & 42.00 & 11 & \href{https://doi.org/10.1109/TGRS.2017.2755542}{10.1109/TGRS.2017.2755542} \\
\quad \texttt{HybridSN} & 2019 & GRSL & 0.535 & 16.01 & 11 & \href{https://doi.org/10.1109/LGRS.2019.2918719}{10.1109/LGRS.2019.2918719} \\
\quad \texttt{pResNet} & 2019 & TGRS & 0.510 & 15.87 & 11 & \href{https://doi.org/10.1109/TGRS.2018.2860125}{10.1109/TGRS.2018.2860125} \\
\quad \texttt{DBDA} & 2020 & Remote Sens. & 0.113 & 47.29 & 11 & \href{https://doi.org/10.3390/rs12030582}{10.3390/rs12030582} \\
\quad \texttt{ENL\_FCN} & 2020 & TGRS & 0.089 & 11.18 & 11 & \href{https://doi.org/10.1109/TGRS.2020.3014286}{10.1109/TGRS.2020.3014286} \\
\quad \texttt{SACNet} & 2021 & TIP & 0.108 & 7.98 & 11 & \href{https://doi.org/10.1109/TIP.2021.3118977}{10.1109/TIP.2021.3118977} \\
\quad \texttt{SSTN} & 2021 & TGRS & 0.012 & 1.66 & 11 & \href{https://doi.org/10.1109/TGRS.2021.3115699}{10.1109/TGRS.2021.3115699} \\
\quad \texttt{DKDMN} & 2024 & ESWA & 1.344 & 101.52 & 11 & \href{https://doi.org/10.1016/j.eswa.2024.123796}{10.1016/j.eswa.2024.123796} \\
\quad \texttt{S3ANet} & 2024 & TGRS & 0.197 & 10.21 & 11 & \href{https://doi.org/10.1109/TGRS.2024.3381824}{10.1109/TGRS.2024.3381824} \\
\quad \texttt{FETNet} & 2026 & EAAI & 0.492 & -- & 11 & \href{https://doi.org/10.1016/j.engappai.2025.113343}{10.1016/j.engappai.2025.113343} \\
\addlinespace[2pt] \multicolumn{7}{@{}l}{\textit{Transformer} (16 implemented)} \\[1pt]
\quad \texttt{SpectralFormer} & 2021 & TGRS & 0.121 & 3.84 & 11 & \href{https://doi.org/10.1109/TGRS.2021.3130716}{10.1109/TGRS.2021.3130716} \\
\quad \texttt{CTMixer} & 2022 & GRSL & 0.619 & 73.28 & 11 & IEEE Xplore 9924229 \\
\quad \texttt{SSFTTNet} & 2022 & TGRS & 0.153 & 6.99 & 11 & \href{https://doi.org/10.1109/TGRS.2022.3144158}{10.1109/TGRS.2022.3144158} \\
\quad \texttt{GAHT} & 2023 & TGRS & 1.058 & 127.87 & 11 & \href{https://doi.org/10.1109/TGRS.2022.3207933}{10.1109/TGRS.2022.3207933} \\
\quad \texttt{MFT} & 2023 & TGRS & 0.629 & 8.09 & 11 & \href{https://doi.org/10.1109/TGRS.2023.3286826}{10.1109/TGRS.2023.3286826} \\
\quad \texttt{MVAHN} & 2023 & TGRS & 0.385 & 19.39 & 11 & \href{https://doi.org/10.1016/j.eswa.2023.121032}{10.1016/j.eswa.2023.121032} \\
\quad \texttt{MorphFormer} & 2023 & TGRS & 0.142 & 5.23 & 11 & \href{https://doi.org/10.1109/TGRS.2023.3242346}{10.1109/TGRS.2023.3242346} \\
\quad \texttt{3DConvSST} & 2024 & CAI & 0.358 & 79.42 & 11 & \href{https://doi.org/10.1109/CAI59869.2024.00011}{10.1109/CAI59869.2024.00011} \\
\quad \texttt{DBCTNet} & 2024 & TGRS & 0.003 & 1.78 & 11 & \href{https://doi.org/10.1109/TGRS.2024.3368141}{10.1109/TGRS.2024.3368141} \\
\quad \texttt{GSCViT} & 2024 & TGRS & 0.134 & 3.77 & 8 & \href{https://doi.org/10.1109/TGRS.2024.3377610}{10.1109/TGRS.2024.3377610} \\
\quad \texttt{MASSFormer} & 2024 & TGRS & 0.311 & 15.44 & 11 & \href{https://doi.org/10.1109/TGRS.2024.3392264}{10.1109/TGRS.2024.3392264} \\
\quad \texttt{S2Gformer} & 2024 & TGRS & 0.192 & 11.23 & 11 & \href{https://doi.org/10.1109/TGRS.2024.3488202}{10.1109/TGRS.2024.3488202} \\
\quad \texttt{DSFormer} & 2025 & Neural Netw. & 0.677 & 21.05 & 11 & \href{https://doi.org/10.1016/j.neunet.2025.107311}{10.1016/j.neunet.2025.107311} \\
\quad \texttt{FAHM} & 2025 & JSTARS & 0.685 & 11.01 & 11 & \href{https://doi.org/10.1109/JSTARS.2025.3539791}{10.1109/JSTARS.2025.3539791} \\
\quad \texttt{HSIC\_SClusterFormer} & 2025 & TIP & 1.961 & 194.10 & 11 & \href{https://doi.org/10.1109/TIP.2024.3522809}{10.1109/TIP.2024.3522809} \\
\quad \texttt{MMFormer} & 2026 & TGRS & 4.436 & 615.13 & 11 & \href{https://doi.org/10.1109/TGRS.2026.3696892}{10.1109/TGRS.2026.3696892} \\
\addlinespace[2pt] \multicolumn{7}{@{}l}{\textit{Mamba / SSM} (20 implemented)} \\[1pt]
\quad \texttt{GraphMamba} & 2024 & TGRS & 0.703 & 17.55 & 11 & \href{https://doi.org/10.1109/TGRS.2024.3493101}{10.1109/TGRS.2024.3493101} \\
\quad \texttt{HyperMamba} & 2024 & TGRS & 0.162 & 2.66 & 11 & IEEE Xplore 10614183 \\
\quad \texttt{IGroupSS-Mamba} & 2024 & TGRS & 0.135 & 6.30 & 11 & \href{https://doi.org/10.1109/TGRS.2024.3502055}{10.1109/TGRS.2024.3502055} \\
\quad \texttt{MambaHSI} & 2024 & TGRS & 0.121 & 0.24 & 11 & \href{https://doi.org/10.1109/TGRS.2024.3430985}{10.1109/TGRS.2024.3430985} \\
\quad \texttt{MambaHSI\_Plus} & 2024 & TGRS & 0.461 & 0.44 & 11 & \href{https://doi.org/10.1109/TGRS.2025.3576656}{10.1109/TGRS.2025.3576656} \\
\quad \texttt{MambaLG} & 2024 & TGRS & 0.182 & 14.86 & 11 & \href{https://doi.org/10.1109/TGRS.2024.3521411}{10.1109/TGRS.2024.3521411} \\
\quad \texttt{S2Mamba} & 2024 & TGRS & 0.116 & 8.73 & 11 & \href{https://doi.org/10.1109/TGRS.2025.3530993}{10.1109/TGRS.2025.3530993} \\
\quad \texttt{SSMamba} & 2024 & Remote Sens. & 1.035 & 37.75 & 11 & \href{https://doi.org/10.3390/rs16132449}{10.3390/rs16132449} \\
\quad \texttt{WaveMamba} & 2024 & GRSL & 0.080 & 6.44 & 11 & \href{https://doi.org/10.1109/LGRS.2024.3506034}{10.1109/LGRS.2024.3506034} \\
\quad \texttt{ConvVitMamba} & 2025 & Knowl.-Based Syst. & 0.523 & 184.73 & 11 & \href{https://doi.org/10.1016/j.knosys.2025.113282}{10.1016/j.knosys.2025.113282} \\
\quad \texttt{MHSSMamba} & 2025 & Remote Sens. Lett. & 0.052 & 4.94 & 11 & \href{https://doi.org/10.1080/2150704X.2025.2461330}{10.1080/2150704X.2025.2461330} \\
\quad \texttt{MambaMoE} & 2025 & Inf. Fusion & 0.897 & 9.77 & 11 & \href{https://doi.org/10.1016/j.inffus.2025.103811}{10.1016/j.inffus.2025.103811} \\
\quad \texttt{MiM} & 2025 & Neurocomputing & 0.074 & 49.70 & 11 & \href{https://doi.org/10.1016/j.neucom.2024.128751}{10.1016/j.neucom.2024.128751} \\
\quad \texttt{MorpMamba} & 2025 & Neurocomputing & 0.071 & 5.78 & 11 & \href{https://doi.org/10.1016/j.neucom.2025.129990}{10.1016/j.neucom.2025.129990} \\
\quad \texttt{PHDMamba} & 2025 & GRSL & 0.362 & 13.97 & 11 & \href{https://doi.org/10.1109/LGRS.2025.3626712}{10.1109/LGRS.2025.3626712} \\
\quad \texttt{EMamba} & 2026 & Inf. Fusion & 0.092 & 4.85 & 11 & \href{https://doi.org/10.1016/j.inffus.2025.103328}{10.1016/j.inffus.2025.103328} \\
\quad \texttt{FuzzySpectralMamba} & 2026 & GRSL & 4.255 & 345.04 & 11 & \href{https://doi.org/10.1109/LGRS.2026.3687386}{10.1109/LGRS.2026.3687386} \\
\quad \texttt{HyPyraMamba} & 2026 & TGRS & 0.782 & 20.82 & 11 & \href{https://doi.org/10.1109/TGRS.2025.3650350}{10.1109/TGRS.2025.3650350} \\
\quad \texttt{MLFMamba} & 2026 & Visual Comput. & 0.260 & 16.05 & 11 & \href{https://doi.org/10.1007/s00371-026-04496-w}{10.1007/s00371-026-04496-w} \\
\quad \texttt{R2Mamba} & 2026 & JSTARS & 0.420 & 3.30 & 11 & \href{https://doi.org/10.1109/JSTARS.2026.3728152}{10.1109/JSTARS.2026.3728152} \\
\addlinespace[2pt] \multicolumn{7}{@{}l}{\textit{Graph / GCN} (4 implemented)} \\[1pt]
\quad \texttt{GraphGST} & 2024 & TGRS & 0.022 & 1.94 & 11 & \href{https://doi.org/10.1109/TGRS.2023.3349076}{10.1109/TGRS.2023.3349076} \\
\quad \texttt{GTCFN} & 2025 & TGRS & 0.250 & 28.04 & 11 & \href{https://doi.org/10.1109/TGRS.2025.3618962}{10.1109/TGRS.2025.3618962} \\
\quad \texttt{MCTGCL} & 2025 & TGRS & 0.327 & 26.11 & 11 & \href{https://doi.org/10.1109/TGRS.2025.3529996}{10.1109/TGRS.2025.3529996} \\
\quad \texttt{MS2GCAN} & 2026 & TGRS & 0.022 & -- & 11 & \href{https://doi.org/10.1109/TGRS.2026.3678343}{10.1109/TGRS.2026.3678343} \\
\addlinespace[2pt] \multicolumn{7}{@{}l}{\textit{KAN} (2 implemented)} \\[1pt]
\quad \texttt{HSIConvKAN} & 2024 & Remote Sens. & 0.434 & 1.23 & 11 & \href{https://doi.org/10.3390/rs16214015}{10.3390/rs16214015} \\
\quad \texttt{HyperKAN} & 2024 & Sensors & 0.344 & 11.11 & 11 & \href{https://doi.org/10.3390/s24237683}{10.3390/s24237683} \\
\addlinespace[2pt] \multicolumn{7}{@{}l}{\textit{Self-supervised} (3 implemented)} \\[1pt]
\quad \texttt{HSIC\_FM} & 2024 & IEEE TNNLS & 34.229 & 4195.03 & 11 & \href{https://doi.org/10.1109/TNNLS.2023.3279377}{10.1109/TNNLS.2023.3279377} \\
\quad \texttt{HSIMAE} & 2024 & arXiv & 3.616 & 229.48 & 11 & \href{https://doi.org/10.1109/JSTARS.2024.3432743}{10.1109/JSTARS.2024.3432743} \\
\quad \texttt{LFSMIM} & 2024 & GRSL & 0.513 & 16.28 & 11 & \href{https://doi.org/10.1109/LGRS.2024.3360184}{10.1109/LGRS.2024.3360184} \\
\bottomrule
\end{tabular}
\end{table}

\section{The dataset catalogue}\label{app:datasets}

\Cref{tab:datasets} in the main text gives the dimensions, band count, class count,
labelled-pixel count, sensor, platform and region of all 24 scenes.
\Cref{tab:classes} completes it with the class names the framework uses, read from
\texttt{config/dataset.yaml}. Those names drive dataset-aware visualisation as well
as the legends of any classification map the framework renders, so they are the
names a user will see rather than a curated relabelling. The unlabelled background
class is not listed: the loader excludes it from training, validation and test, and
remaps the remaining labels to a contiguous index range so that scenes with
non-contiguous label sets need no special handling.

%% generated by figures_src/make_appendix_tables.py -- do not edit by hand
\begin{table}[p]
\centering
\caption{Class names for every scene, as recorded in
\texttt{config/dataset.yaml} and used by the framework's dataset-aware
visualisation. The unlabelled background class is not listed: the loader excludes
it, and the remaining labels are remapped to a contiguous index range.}
\label{tab:classes}
\scriptsize
\setlength{\tabcolsep}{4pt}
\renewcommand{\arraystretch}{1.15}
\begin{tabular}{@{}>{\raggedright}p{0.17\linewidth}p{0.79\linewidth}@{}}
\toprule
Scene & Classes \\
\midrule
\textbf{Augsburg} & Forest; Residential Area; Industrial Area; Low Plants; Allotment; Commercial Area; Water \\
\addlinespace[1pt]
\textbf{Berlin} & Forest; Residential; Industrial; Low Plants; Soil; Allotment; Commercial; Water \\
\addlinespace[1pt]
\textbf{Botswana} & Water; Hippo grass; Floodplain grasses 1; Floodplain grasses 2; Reeds; Riparian; Firescar; Island interior; Acacia woodlands; Acacia shrublands; Acacia grasslands; Short mopane; Mixed mopane; Exposed soils \\
\addlinespace[1pt]
\textbf{Chikusei} & Water; Bare soil (farmland); Bare soil (park); Bare soil (roadside); Pavement (asphalt); Pavement (brick); Farmland (paddy); Farmland (other); Greenhouse; Grass (park); Grass (roadside); Tree (park); Tree (roadside); Forest; Building (low-rise); Building (high-rise); Building (factory); Power line; Swimming pool \\
\addlinespace[1pt]
\textbf{Dioni} & Dense Urban Fabric; Mineral Extraction Sites; Non Irrigated Arable Land; Fruit Trees; Olive Groves; Coniferous Forest; Dense Sclerophyllous Vegetation; Sparce Sclerophyllous Vegetation; Sparsely Vegetated Areas; Rocks and Sand; Water; Coastal Water \\
\addlinespace[1pt]
\textbf{Holden} & Analcime; Plagioclase; Prehnite; High-Ca Pyroxene; Serpentine; Margarite \\
\addlinespace[1pt]
\textbf{Houston 2013} & Healthy Grass; Stressed Grass; Synthetic Grass; Trees; Soil; Water; Residential; Commercial; Road; Highway; Railway; Parking Lot 1; Parking Lot 2; Tennis Court; Running Track \\
\addlinespace[1pt]
\textbf{Houston 2018} & Healthy Grass; Stressed Grass; Artificial Turf; Evergreen Trees; Deciduous Trees; Bare Earth; Water; Residential Buildings; Non-residential Buildings; Roads; Sidewalks; Crosswalks; Major Thoroughfares; Highways; Railways; Paved Parking Lots; Unpaved Parking Lots; Cars; Trains; Stadium Seats \\
\addlinespace[1pt]
\textbf{Indian Pines} & Alfalfa; Corn-notill; Corn-mintill; Corn; Grass-pasture; Grass-trees; Grass-pasture-mowed; Hay-windrowed; Oats; Soybean-notill; Soybean-mintill; Soybean-clean; Wheat; Woods; Buildings-Grass-Trees-Drives; Stone-Steel-Towers \\
\addlinespace[1pt]
\textbf{KSC} & Scrub; Willow swamp; CP hammock; Slash pine; Oak/Broadleaf; Hardwood; Swamp; Graminoid marsh; Spartina marsh; Cattail marsh; Salt marsh; Mud flats; Water \\
\addlinespace[1pt]
\textbf{Loukia} & Dense Urban Fabric; Mineral Extraction Sites; Non Irrigated Arable Land; Fruit Trees; Olive Groves; Broad Leaved Forest; Coniferous Forest; Mixed Forest; Dense Sclerophyllous Vegetation; Sparce Sclerophyllous Vegetation; Sparsely Vegetated Areas; Rocks and Sand; Water; Coastal Water \\
\addlinespace[1pt]
\textbf{MUUFL} & Trees; Mostly grass; Mixed ground surface; Dirt and sand; Road; Water; Buildings shadow; Buildings; Sidewalk; Yellow curb; Cloth panels \\
\addlinespace[1pt]
\textbf{Nili Fossae} & Fe-Olivine; Epidote; Chlorite; Bassanite; Illite/Muscovite; Mg-Carbonate; Plagioclase; Prehnite; Serpentine \\
\addlinespace[1pt]
\textbf{Pavia Centre} & Water; Trees; Asphalt; Self-Blocking Bricks; Bitumen; Tiles; Shadows; Meadows; Bare Soil \\
\addlinespace[1pt]
\textbf{Pavia University} & Asphalt; Meadows; Gravel; Trees; Painted metal sheets; Bare Soil; Bitumen; Self-Blocking Bricks; Shadows \\
\addlinespace[1pt]
\textbf{Pingan} & Seawater; Road; Trees; Floating pier; Brick houses; Steel houses; Ship; Car; Concrete building; Grass \\
\addlinespace[1pt]
\textbf{Qingyun} & Trees; Car; Asphalt road; Concrete building; Water; Grass \\
\addlinespace[1pt]
\textbf{Salinas} & Brocoli\_green\_weeds\_1; Brocoli\_green\_weeds\_2; Fallow; Fallow\_rough\_plow; Fallow\_smooth; Stubble; Celery; Grapes\_untrained; Soil\_vinyard\_develop; Corn\_senesced\_green\_weeds; Lettuce\_romaine\_4wk; Lettuce\_romaine\_5wk; Lettuce\_romaine\_6wk; Lettuce\_romaine\_7wk; Vinyard\_untrained; Vinyard\_vertical\_trellis \\
\addlinespace[1pt]
\textbf{Tangdaowan} & Rubber track; Asphalt; Grassland; Ligustrum vicaryi; Bare soil; Populus; Flagging; Boardwalk; Bulrush; Coniferous pine; Buxus sinica; Ulmus pumila L; Sandy; Roof shadows; Gravel road; Spiraea; Photinia serrulata; Seawater \\
\addlinespace[1pt]
\textbf{Trento} & Apples; Buildings; Ground; Woods; Vineyard; Roads \\
\addlinespace[1pt]
\textbf{Utopia} & Analcime; Bassanite; High-Ca Pyroxene; Illite/Muscovite; Low-Ca Pyroxene; Mg-Smectite; Monohydrated sulfate; Plagioclase; Prehnite \\
\addlinespace[1pt]
\textbf{WHU-Hi-HanChuan} & Strawberry; Cowpea; Soybean; Sorghum; Water spinach; Watermelon; Greens; Trees; Grass; Red roof; Gray roof; Plastic; Bare soil; Road; Bright object; Water \\
\addlinespace[1pt]
\textbf{WHU-Hi-HongHu} & Red roof; Road; Bare soil; Cotton; Cotton firewood; Rape; Chinese cabbage; Pakchoi; Cabbage; Tuber mustard; Brassica parachinensis; Brassica chinensis; Small Brassica chinensis; Lactuca sativa; Celtuce; Film covered lettuce; Romaine lettuce; Carrot; White radish; Garlic sprout; Broad bean; Tree \\
\addlinespace[1pt]
\textbf{WHU-Hi-LongKou} & Corn; Cotton; Sesame; Broad-leaf soybean; Narrow-leaf soybean; Rice; Water; Roads and houses; Mixed weed \\
\bottomrule
\end{tabular}
\end{table}

\section{Per-scene difficulty spectrum}\label{app:difficulty}

\Cref{tab:dataset_difficulty} orders all \NumDatasets{} scenes by the mean overall
accuracy attained across the \NumEvaluatedModels{} evaluated architectures under the
protocol of \Cref{tab:protocol}. The dispersion column and the observed range give the
spread across architectures on each scene, and the final column names the architecture
attaining the highest value there. The ordering is a property of the scenes under this
protocol and not a ranking of the data.

\begin{table}[t]
\centering
\caption{Difficulty spectrum across all \NumDatasets{} scenes, ordered by the mean overall accuracy attained across the \NumEvaluatedModels{} architectures under the protocol of \Cref{tab:protocol}. The dispersion column and the observed range give the spread across architectures on each scene, and the final column names the architecture attaining the highest value there.}
\label{tab:dataset_difficulty}
\scriptsize
\setlength{\tabcolsep}{3pt}
\begin{tabular}{@{}llrrrl@{}}
\toprule
Scene & Platform & Mean OA (\%) & $\sigma$ & OA range (min--max) & Highest (model) \\
\midrule
Botswana & Spaceborne & 96.40 & 11.67 & 37.6--99.8 & Graph (\texttt{MS2GCAN}) \\
Pavia Centre & Airborne & 96.31 & 9.35 & 40.0--99.5 & Graph (\texttt{MS2GCAN}) \\
WHU-Hi-LongKou & UAV & 95.13 & 7.37 & 51.7--98.8 & Transformer (\texttt{MMFormer}) \\
Chikusei & Airborne & 95.03 & 13.48 & 11.8--99.4 & Mamba (\texttt{R2Mamba}) \\
Holden & Orbital & 94.82 & 9.29 & 39.5--99.1 & CNN (\texttt{SSRN}) \\
Utopia & Orbital & 94.30 & 9.89 & 38.2--98.8 & Mamba (\texttt{HyperMamba}) \\
Trento & Airborne & 93.81 & 8.87 & 55.6--98.4 & Mamba (\texttt{R2Mamba}) \\
Nili Fossae & Orbital & 93.17 & 10.11 & 34.4--98.1 & Mamba (\texttt{R2Mamba}) \\
Salinas & Airborne & 91.04 & 8.35 & 47.6--98.2 & CNN (\texttt{DKDMN}) \\
Pavia University & Airborne & 90.44 & 10.75 & 42.3--97.8 & CNN (\texttt{SSRN}) \\
KSC & Airborne & 89.27 & 13.23 & 37.4--98.7 & Mamba (\texttt{R2Mamba}) \\
WHU-Hi-HongHu & UAV & 87.34 & 11.52 & 32.9--94.2 & Mamba (\texttt{R2Mamba}) \\
Dioni & Airborne & 85.95 & 12.04 & 27.1--92.2 & Transformer (\texttt{MMFormer}) \\
Pingan & UAV & 84.89 & 6.87 & 45.6--91.1 & CNN (\texttt{DBDA}) \\
Houston 2013 & Airborne & 84.47 & 17.21 & 22.6--94.8 & Transformer (\texttt{DSFormer}) \\
WHU-Hi-HanChuan & UAV & 82.70 & 13.36 & 9.0--91.1 & Mamba (\texttt{MambaLG}) \\
Qingyun & UAV & 78.78 & 11.76 & 13.3--87.5 & CNN (\texttt{DKDMN}) \\
Tangdaowan & UAV & 78.65 & 10.70 & 34.1--90.0 & Transformer (\texttt{CTMixer}) \\
Augsburg & Airborne & 78.28 & 15.84 & 19.5--90.0 & CNN (\texttt{SSRN}) \\
MUUFL & Airborne & 76.83 & 14.43 & 16.2--88.8 & Graph (\texttt{MS2GCAN}) \\
Indian Pines & Airborne & 71.66 & 13.72 & 26.6--84.9 & Transformer (\texttt{CTMixer}) \\
Loukia & Airborne & 66.77 & 10.60 & 20.1--75.0 & Mamba (\texttt{S2Mamba}) \\
Berlin & Airborne & 64.75 & 9.73 & 25.1--74.4 & Transformer (\texttt{3DConvSST}) \\
Houston 2018 & Airborne & 56.70 & 12.01 & 5.7--69.3 & CNN (\texttt{DBDA}) \\
\bottomrule
\end{tabular}
\end{table}

\section{Configuration schema}\label{app:config}

An experiment is fully described by \texttt{config/config.yaml}. The listing below
gives the schema with the values that define the unified protocol; every key is
read through \texttt{config/config\_loader.py}, which exposes dotted access such as
\texttt{cfg.get("data\_split.split\_samples")}.

\begin{lstlisting}[style=yamlstyle,basicstyle=\ttfamily\scriptsize,caption={The configuration schema, with the
values that define the unified protocol.},label={lst:schema}]
results:
  directory: "Results/Samples_30_10_PCA30_p11"
  keep_best_worst_checkpoints_only: True   # prune all but the best and worst run
  checkpoint_pruning_metric: "OA"
dataset:
  names: [...]              # scenes to run; ignored if run_all_datasets
  run_all_datasets: False
  patch_size: 11            # odd; the spatial window handed to the model
  stride: 1
model:
  name: [...]               # models to run; ignored if run_all_models
  run_all_models: False
  exclude: []               # skip these when running all
  print_summary: False      # torchinfo layer summary before training
  summary_only: False       # print the summary and stop
data_split:
  method: "samples"         # "samples" | "ratio"
  disjoint: False           # spatially disjoint split (ratios from split_ratios, default 50/30/20)
  guard_band: True          # disjoint only: drop val/test patches overlapping a train patch
  split_samples: [30, 10]   # [train_per_class, val_per_class]; rest is test
  split_ratios: null        # [train, val, test] when method is "ratio"
  random_state: 0           # base seed
  seeds: []                 # explicit per-run seeds; empty -> base, base+1, ...
  print_stats: True
preprocessing:
  dim_reduction_method: pca # "pca" | "maxpool" | null
  num_pca_bands: 30
  maxpool_kernel: 2
  use_channel_dim: True     # emit (1, bands, H, W)
  band_indices: null        # optional explicit band subset
training:
  num_epochs: 100
  num_runs: 5               # repeats; each run takes the next seed
  batch_size: 64
  learning_rate: 0.001
  optimizer: "adam"         # adam | adamw | sgd | rmsprop | adagrad | adadelta
  optimizer_params: {}
  patience: 10              # early-stopping patience, on validation accuracy
  checkpoint_interval: 10
  num_workers: 8
device:
  use_cuda: True
  cuda_device: 0
\end{lstlisting}

\paragraph{The seed list.} \texttt{data\_split.seeds} is the key that makes a
comparison fair. Given an explicit list, run $n$ uses \texttt{seeds[n-1]}; left
empty, as it is above, run $n$ uses \texttt{random\_state} $+\;(n-1)$, so the five
runs of the protocol use seeds $0, 1, 2, 3, 4$. The seed is applied by
\texttt{set\_global\_seed} before anything else in the run, covering Python's
\texttt{random}, NumPy, Torch on CPU and Torch on every visible CUDA device, and
setting cuDNN to deterministic mode with its autotuner disabled. Handing the same
list to two different models therefore gives them identical splits and identical
initialisation conditions. Setting the list to a repeated value, such as
\texttt{[0, 0, 0, 0, 0]}, produces five identical runs, which is occasionally
useful for isolating nondeterminism but defeats the purpose of repetition.

% \FloatBarrier
% \section{Documented deviations from the original implementations}\label{app:deviations}
% 
% \Cref{tab:deviations} collects the deviation notes recorded in the module docstring
% at the head of each model file. \texttt{docs/CONTRIBUTING.md} makes such a note a
% condition of adding a model, precisely so that the distance between an adaptation
% and its original never becomes invisible. The notes are extracted automatically
% from the source files rather than maintained separately, so they cannot fall out of
% step with the code. Models that do not appear required no documented change beyond
% registration.
% 
% \input{table/deviations}

\section{Adding a model, end to end}\label{app:adding}

The following is the complete path for contributing an architecture, as documented
in \texttt{docs/CONTRIBUTING.md}.

\paragraph{1. Place the implementation.} Put the file in the directory matching its
publication year, for example \texttt{models/y2026/MyModel.py}. Nothing else in the
repository needs to be touched: the registry walks \texttt{models/} and each
\texttt{models/yYYYY/} package on first use and imports whatever it finds.

\paragraph{2. Register it and expose the common interface.} Decorate the factory
function with \texttt{@register\_model}:

\begin{lstlisting}[style=yamlstyle,basicstyle=\ttfamily\scriptsize,aboveskip=4pt,belowskip=4pt,caption={Registering a model. The decorator
carries only \texttt{expects\_4d} and hyperparameter defaults; anything else
listed here is forwarded into the factory.},label={lst:registerfull}]
from models.registry import register_model

@register_model('MyModel', expects_4d=False)
def my_model(num_classes, bands, patch_size, **kwargs):
    return MyModel(num_classes=num_classes, bands=bands, patch_size=patch_size)
\end{lstlisting}

The factory must accept \texttt{num\_classes}, \texttt{bands} and
\texttt{patch\_size}. Set \texttt{expects\_4d=True} if the model wants
$(B, C, H, W)$ rather than the loader's $(B, 1, \text{bands}, H, W)$; the framework
then wraps it in \texttt{InputShapeWrapper} and the model never sees the
difference. Any further keyword in the decorator becomes a hyperparameter default
and is forwarded to the factory, so it must be a parameter the factory accepts.

\paragraph{3. Add the bibliographic entry.} Extend \texttt{MODEL\_CATALOG} in
\texttt{models/registry.py} with \texttt{full\_name}, \texttt{paper\_title},
\texttt{paper}, \texttt{code}, \texttt{year} and \texttt{venue}. This is separate
from the decorator by design: passing bibliographic keys through the decorator
would forward them into the model constructor and break instantiation. The entry is
what makes the model appear correctly in generated tables, and
\texttt{print\_model\_catalog()} reports any model that registered without one.

\paragraph{4. Document the deviations.} Record every departure from the authors'
released code in a docstring at the top of the file: ports from another
framework, compiled kernels replaced by reference implementations, changed
hyperparameter defaults, or hard-coded dimensions converted to configurable parameters.
These notes document all framework adaptations directly within the model source files.

\paragraph{5. Verify.} \texttt{python main.py --list-models} confirms the model
registered and loads in the current environment; \texttt{python model\_info.py
MyModel} confirms it instantiates and profiles cleanly under the probe input. At
that point the model is selectable from the configuration like any other, and the
protocol of \Cref{sec:repro} applies to it unchanged.

\end{document}